\documentclass[journal]{IEEEtran}

\usepackage[utf8]{inputenc} 
\usepackage[T1]{fontenc}    

\usepackage{hyperref}       
\usepackage{url}            
\usepackage{booktabs}       
\usepackage{amsmath,amssymb,amsbsy,amsfonts,amsthm,units,bm}
\usepackage{nicefrac}       
\usepackage{microtype}      
\usepackage{xcolor}         
\usepackage{graphicx}
\usepackage{tikz}
\usepackage{stackengine}
\let\labelindent\relax
\usepackage{enumitem}
\usepackage{tabularx}
\newcolumntype{Y}{>{\centering\arraybackslash}X}

\usepackage{makecell}
\usepackage{indentfirst}
\usepackage{bbm}
\usepackage{verbatim}
\usepackage{colortbl}
\usepackage{marvosym}
\usepackage[most]{tcolorbox} 

\usepackage{tcolorbox}

\usepackage{algorithm,algorithmic}
\usepackage{multirow}
\usepackage{subfiles}

\newtheorem{assumption}{Assumption}

\newcommand{\partitle}[1]{\smallskip \noindent \textbf{#1.}}

\begin{document}

\title{Towards Mitigating Excessive Forgetting in \\LLM Unlearning via Entanglement-Guidance with Proxy Constraint}

\author{
Zhihao~Liu,
Jian~Lou,
Yuke~Hu\textsuperscript{\Letter},
Xiaochen~Li,
Yitian~Cheng,
Tailun~Chen,
Zhizhen~Qin,\\
Zhan~Qin\textsuperscript{\Letter},
Kui~Ren~\IEEEmembership{Fellow,~IEEE}%
\thanks{
Zhihao~Liu, Yuke~Hu, Yitian~Cheng, Tailun~Chen, Zhan~Qin and Kui~Ren are with the Department of Computer Science and Technology, Zhejiang University, Hangzhou 310027, China. Jian Lou is with Sun Yat-sen University. Xiaochen~Li is with UNC Greensboro. Zhizhen Qin is with Amazon, but this work is independent of and outside of the work at Amazon. E-mails:(zhihao\_liu, yukehu, yitian111, tailun.chen, qinzhan, kuiren)@zju.edu.cn, jian.lou@hoiying.net, X\_LI12@uncg.edu, zhizhenq@amazon.com}

\thanks{\textsuperscript{\Letter}Corresponding authors.}%
}

\maketitle

\begin{abstract}

Large language models (LLMs) are trained on massive datasets that may include private or copyrighted content. Due to growing privacy and ownership concerns, data owners may request the removal of their data from trained models. Machine unlearning provides a practical solution by removing the influence of specific data without full retraining. However, most existing methods still suffer from over-unlearning due to the lack of a principled mechanism to regulate the forgetting boundary, leading to unnecessary utility degradation and heightened privacy and robustness risks.

In this work, we propose EGUP (Entanglement-Guided Unlearning with Proxy Constraint), a novel framework that leverages entanglement and proxy constraint to guide the unlearning process while mitigating over-unlearning. Within each iteration, EGUP employs inter-sample entanglement to adaptively reweight the unlearning strength, assigning greater unlearning efforts to forget samples that are semantically closer to retained knowledge. Across iterations, EGUP leverages intra-sample entanglement to track the representation shift of each forget sample and dynamically adjust its unlearning effort. In addition, we incorporate a proxy constraint that approximates the model's expected outputs after unlearning, forming a reference boundary to softly regularize the unlearning process. EGUP is compatible with existing gradient-based objectives and serves as a plug-and-play enhancement. Experiments on TOFU and MUSE show consistent improvements in the unlearning--utility trade-off across multiple LLMs. Moreover, EGUP achieves performance close to the retrained model while remaining scalable and robust.

\end{abstract}

\section{Introduction}

Large language models (LLMs) \cite{touvron2023llama,achiam2023gpt,bai2023qwen} demonstrate an extraordinary capacity to absorb and retain knowledge from massive web-scale datasets. However, this memorization capability also brings accompanying privacy risks. Real-world training data often contains private, sensitive, or copyrighted content \cite{patil2023can,karamolegkou2023copyright,liu2026dpzo}, which may unintentionally resurface at inference time or be maliciously extracted by adversaries through targeted attacks such as training data extraction and membership inference \cite{meng2025rr,hu2025vlmmia,hu2023fkge}. Such disclosures threaten individual privacy rights \cite{yao2024survey}, undermine intellectual property protections \cite{wei2024evaluating}, and facilitate harmful misuse \cite{liu2023jailbreaking}. Such risks highlight the need for mechanisms that can reliably remove specific training data (i.e., forget data) from trained models. 

Machine unlearning \cite{cao2015towards,bourtoule2021machine} has emerged as a practical alternative that directly updates model parameters to remove the influence of targeted data, without the need for full retraining from scratch. The common paradigm for unlearning is to first formulate a dedicated unlearning objective and then optimize the model iteratively \cite{liu2025rethinking,jia2024soul,yao2024machine,bao2025module,hu2024eraser}. By explicitly guiding the optimization process toward suppressing memorized information, these approaches significantly reduce computational cost while maintaining model functionality on retained knowledge. Consequently, machine unlearning offers a promising balance between effectiveness, efficiency, and utility, making it particularly suitable for LLM scenarios.

Although unlearning substantially reduces the prohibitive cost of full retraining, it still faces the inherent risk of over-unlearning. When parameter updates become overly aggressive, the model may not only erase the targeted knowledge but also inadvertently degrade useful capabilities, thereby disrupting the desired balance between forgetting and utility \cite{zhao2025rethinking}. In addition, state-of-the-art LLMs are carefully aligned with safety guidelines \cite{miao2026safe}; excessive perturbations on model parameters may weaken these safety guardrails, potentially compromising model alignment and increasing the likelihood of generating unsafe content \cite{lucki2024adversarial,chen-etal-2025-safeeraser,liu2026aligned}. Beyond utility and safety concerns, recent studies further reveal that over-unlearning can amplify privacy and robustness risks. Excessive representation drift may destabilize decision boundaries and induce abnormal confidence behavior, which in turn increases the success rate of membership inference attacks on both forget and retain samples, while simultaneously reducing robustness under adversarial perturbations \cite{hayes2025inexact,xue2025dual}. These findings highlight that over-unlearning is not merely a degradation in performance, but a broader threat to model reliability, safety, and security.

Nevertheless, existing unlearning methods remain inadequate in effectively preventing over-unlearning. Approaches that consider only individual sample properties \cite{huang2024unified,yang2025exploring} fail to account for the complex interactions between forget samples and the retain dataset, which exert significant influence on the forgetting behavior and are crucial for maintaining a desirable balance between effectiveness and utility preservation. Another line of work leverages memorization scores to guide unlearning \cite{zhao2024makes}. However, its reliance on repeated retraining leads to significant computational overhead, making it impractical for LLMs. Furthermore, current unlearning objectives inherently lack reliable termination conditions \cite{liu2025rethinking,ye2025towards} and explicit stopping criteria, leaving no principled guidance on when sufficient unlearning has been achieved.

Despite these substantial efforts in mitigating over-unlearning, effectively preventing it remains a fundamental and largely unsolved challenge in LLM unlearning. We attribute such risks to two main factors. First, forget samples exhibit strong heterogeneity in memorization and their semantic entanglement with the retain dataset. As a result, applying uniform unlearning effort may cause some samples to be over-unlearned while others remain under-unlearned. Second, most existing work lack reliable convergence guarantees and explicit stopping criteria, making it difficult to determine when adequate unlearning has been achieved, thereby increasing the risk of over-unlearning across all forget samples.

In this work, we propose EGUP (\textbf{E}ntanglement \textbf{G}uided \textbf{U}nlearning with \textbf{P}roxy Constraint), an entanglement-guided unlearning framework for large language models that mitigates over-unlearning through entanglement guidance and proxy regularization. EGUP explicitly captures inter- and intra-sample entanglement to guide the unlearning process. Specifically, inter-sample entanglement, defined across samples, captures the semantic coupling between forget and retain samples in the embedding space and leverages this relationship to adaptively modulate unlearning effort across samples via sample-level loss reweighting, enabling more precise and targeted unlearning. In contrast, intra-sample entanglement, defined across iterations, measures the temporal consistency of a forgotten sample’s representations throughout the training process. It also reflects how robust this sample remains to perturbations introduced by updates on other forget samples within the shared parameter space. By serving as a temporal regulation signal, it helps smooth the unlearning trajectory, thereby preventing instability and avoiding overly aggressive unlearning.

Furthermore, the absence of a clear stopping criterion makes it difficult to determine when unlearning is sufficient, increasing the risk of over-unlearning. A naive approach is to use a fixed threshold to terminate unlearning; however, this is unreliable because the optimal stopping point is highly sensitive to variations in model architecture, data distribution, and individual samples. To better address this, we leverage open-source LLMs to construct proxy constraint by predicting retrained model performance on forget samples. For each sample, the LLM is prompted with a few examples illustrating how the retrained model predicts, and then generates proxy outputs that approximate the retrained model's behavior. These proxy outputs are used in a loss-based penalty mechanism to specify how much the model should still ``remember'' the targeted information. By penalizing deviations from these proxy targets, the mechanism provides principled guidance to constrain unnecessary loss escalation and prevent over-unlearning.

We extensively evaluate EGUP on two representative benchmarks, TOFU \cite{maini2024tofu} and MUSE \cite{shi2024muse}. When integrated into diverse mainstream unlearning pipelines, EGUP consistently outperforms baselines and further achieves almost the same performance as full retraining. More importantly, EGUP consistently outperforms baselines that depend on retain data, while using only the averaged retain embedding. Our contributions are summarized as follows:

\begin{itemize}[leftmargin=*, labelindent=0.2em]
    \item \textbf{Entanglement-Guided Unlearning Mechanism}: We design an entanglement-guided mechanism that jointly leverages inter-sample and intra-sample entanglement to regulate the unlearning process. By reweighting sample-level unlearning effort based on semantic entanglement and incorporating a temporal regularization term sensitive to representation drift, the framework achieves more precise and stable unlearning.
    \item \textbf{Proxy Constraint}: We introduce a soft regularization mechanism based on proxy outputs synthesized via in-context learning (ICL) to approximate a retain model’s behavior on forget samples, providing a principled constraint that prevents over-unlearning.

    \item \textbf{Unified and Plug-and-Play}: EGUP is compatible with multiple gradient-based unlearning objectives and can be applied as a plug-and-play enhancement, improving the unlearning--utility trade-off without costly retraining.
    \item \textbf{Strong Empirical Evidence}: Extensive experiments on TOFU and MUSE show that EGUP achieves superior trade-offs across Phi-1.5, LLaMA2-7B, and ICML-7B. Under a 5\% unlearning setting with NPO+GD, it boosts forget quality from 0.0878 to 0.7934 while maintaining utility, substantially outperforming standard baselines.
\end{itemize}

\section{Background and Related Work}

\subsection{Unlearning for LLMs}

Machine unlearning aims to remove the influence of specific training data from a trained model while preserving overall performance on the retained data. Recent works have introduced gradient-based optimization strategies to achieve this goal. Notably, Gradient Difference (GD) \cite{maini2024tofu} and Negative Preference Optimization (NPO) \cite{zhang2024negative} represent two seminal approaches that guide parameter updates to counteract the influence of forget samples while maintaining model stability. Building upon these methods, subsequent studies refine the underlying loss functions, introducing adaptive weighting and optimization mechanisms to achieve more stable and precise unlearning. Huang et al. \cite{huang2024unified} proposed a Sample-wise Adaptive Coefficient, inversely scaled by empirical loss, to balance gradient ascent on forget samples. It prevents redundant updates to already forgotten data while emphasizing those that remain insufficiently erased. However, as noted in their work, relying solely on empirical loss as a measure of sample importance introduces bias and fails to capture representational complexity. In parallel, Yang et al. \cite{yang2025exploring} identify two complementary criteria for loss reweighting in unlearning: Saturation, which prioritizes under-optimized samples, and Importance, which emphasizes samples most influential for overall loss minimization. Together, these methods demonstrate the potential of adaptive weighting schemes to balance unlearning quality and model retention.

\subsection{Memorization}
Deep neural networks are known to memorize their training data. Feldman et al. \cite{feldman2020does} show that instance-level memorization is not merely an artifact of overfitting but a key mechanism for generalization under long-tailed distributions. Subsequent studies \cite{feldman2020does,brown2021language} further show that moderate memorization helps capture rare or low-frequency patterns critical for generalization.

Beyond its role in generalization, recent research \cite{torkzadehmahani2024improved} reveals that memorization and unlearning are deeply intertwined. At the extreme, non-memorized examples can be regarded as trivially unlearned, since their predictions would remain unchanged without being included in training. In contrast, highly memorized examples are much harder to erase, as their influence is more deeply embedded in the model parameters. Zhao et al. \cite{zhao2024makes} systematically analyze the role of memorization in unlearning difficulty. They find that approximate unlearning methods are generally more successful on forget sets with low-memorization samples, while highly memorized ones are substantially harder to remove. To address this, they propose RUM (Refined-Unlearning Meta-algorithm), which partitions the forget set into homogeneous subsets based on memorization scores, achieving more stable and fine-grained unlearning. Despite its strengths, RUM still faces several limitations: (1) The computation of memorization scores incurs high computational overhead. (2) It focuses only on inter-sample correlations, ignoring intra-sample representational dynamics essential for modeling temporal entanglement and stable unlearning. (3) Their empirical validation is limited to ResNet-based image classification tasks without exploring the generalization to LLMs.

\subsection{Over-unlearning in Unlearning}

Continued gradient updates during unlearning can excessively alter the model’s behavior on forget samples, causing their predictions to diverge from those on naturally unseen inputs. Such over-unlearning not only degrades generalization and model utility but also increases vulnerability to Membership Inference Attacks (MIA) \cite{shokri2017membership,hu2025fkge}, potentially leaking whether a sample in the forget dataset $\mathcal{D}_f$ was part of the original training set $\mathcal{D}$ \cite{shi2024muse,hayes2025inexact,pawelczyk2024machine}.

To address these challenges, Huang et al. \cite{huang2024unified} incorporate gradient upper bounds to stabilize the optimization process and facilitate early convergence. However, it can inadvertently result in under-unlearning of sensitive information. An alternative line of work seeks to reverse the conventional training objective using an assistant LLM. For instance, Ji et al. \cite{ji2024reversing} propose training an auxiliary model with inverted training signals to address both over- and under-unlearning. While promising, this approach comes with significant computational overhead, as training an additional assistant model can be resource-intensive.

\section{Preliminary}

\subsection{Unlearning Problem Formulation}

In machine unlearning, let $\mathcal{D}=\{z_i\}_{i=1}^N$ represent a pretraining dataset of $N$ data points, including $z_i=(x_i,y_i)$ features and labels in supervised learning. The training dataset $\mathcal{D}$ comprises two subsets, namely $\mathcal{D}_f$ and $\mathcal{D}_r$, representing the data points to be unlearned and the retain data points, respectively, wherein $\mathcal{D}_r = \mathcal{D} \setminus \mathcal{D}_f$. Denote $\mathcal{M}_t = \mathcal{M}(\mathcal{D})$ as the model trained on the entire dataset $\mathcal{D}$. The goal of MU is to produce an unlearned model $\mathcal{M}_u$ that closely resembles the model $\mathcal{M}_u^*$ trained solely on $\mathcal{D}_r$, i.e., $\mathcal{M}_u^* = \mathcal{M}(\mathcal{D}_r)$. 



\subsection{Mathematical Modeling for Unlearning}

A mainstream approach in current LLM unlearning research is to fine-tune the pre-trained model with specially designed objectives that suppress the influence of forget data while preserving retained knowledge. Although existing methods differ in their specific formulations, they can be broadly described by the following optimization structure:
\begin{equation}
    \min_{\theta'} L_{\text{unlearn}}(\theta')
    =
    \min_{\theta'}
    \left(
    - L_f(\theta',\mathcal{D}_f)
    +
    \alpha L_r(\theta',\mathcal{D}_r)
    \right),
    \label{equa_gd}
\end{equation}
where $\mathcal{D}_f$ and $\mathcal{D}_r$ denote the forget and retain datasets, respectively, and $\alpha$ controls the strength of retain preservation. The forget term $L_f(\theta',\mathcal{D}_f)$ measures the model's prediction quality on the forget data, while the retain term $L_r(\theta',\mathcal{D}_r)$ measures its prediction quality on the retain data. Therefore, Eq.~\ref{equa_gd} maximizes the loss on $\mathcal{D}_f$ to remove targeted information, while minimizing the loss on $\mathcal{D}_r$ to preserve model utility.

This formulation covers several commonly used gradient-based unlearning objectives, including Gradient Ascent (GA), Gradient Difference (GD), Negative Preference Optimization (NPO), and their combinations. We provide the detailed formulations of these objectives in Appendix~\ref{app:unlearning_objectives}. Our method is built on this general formulation and can be applied as a plug-and-play enhancement to these existing objectives.

\subsection{Unlearning Benchmark}
In large language model unlearning research, evaluating a model’s ability to forget specific data while preserving overall performance is crucial \cite{miao2025towards,wang2026deconstruct}. To standardize such evaluations, several benchmarks have been proposed \cite{maini2024tofu,li2024wmdp,shi2024muse}, among which TOFU \cite{maini2024tofu} and MUSE \cite{shi2024muse} are the most representative. TOFU evaluates the model’s ability to forget specific data using around 200 fictitious author profiles, comparing outputs between the retrained and unlearned models to assess both forget quality and model utility. MUSE provides a comprehensive, multi-dimensional evaluation across six properties, including knowledge memory (KnowMem), verb memory (VerbMem), and privacy leakage (PrivLeak) on large-scale datasets such as books and news articles. These benchmarks together provide a more standardized framework for assessing unlearning methods in large language models, enabling quantitative comparison of unlearning effectiveness and utility preservation.

\subsection{Memorization Score}

Deep neural networks inherently memorize their training data. Feldman et al. \cite{feldman2020does} show that such instance-level memorization is essential for generalization under long-tailed distributions. This perspective underpins the use of memorization scores below as a proxy to quantify the degree to which individual samples are preserved in the model's memory.

\begin{assumption}[Memorization Score\cite{feldman2020does}]
The memorization score for an example $i\in \mathcal{D}$ with respect to a training dataset $\mathcal{D}$ and algorithm $\mathcal{A}$ is defined as:
\begin{small}
    \begin{equation}
\text{mem}(\mathcal{A}, \mathcal{D}, i) = \Pr_{f \sim \mathcal{A}(\mathcal{D})} \left[ f(x_i) = y_i \right] - \Pr_{f \sim \mathcal{A}(\mathcal{D} \setminus {i})} \left[ f(x_i) = y_i \right],
\end{equation}
\end{small}
where $x_i$ and $y_i$ are the feature and label of example $z_i$.
\end{assumption}

Intuitively, the memorization score for example $z_i$ is high if including it in training yields a different prediction distribution on that example than excluding it from training would have, reflecting the model’s reliance on direct exposure to $z_i$.

\section{Limitations of Existing Unlearning Methods and Our Insights}

\begin{figure}[t]
    \centering
    \includegraphics[width=\linewidth]{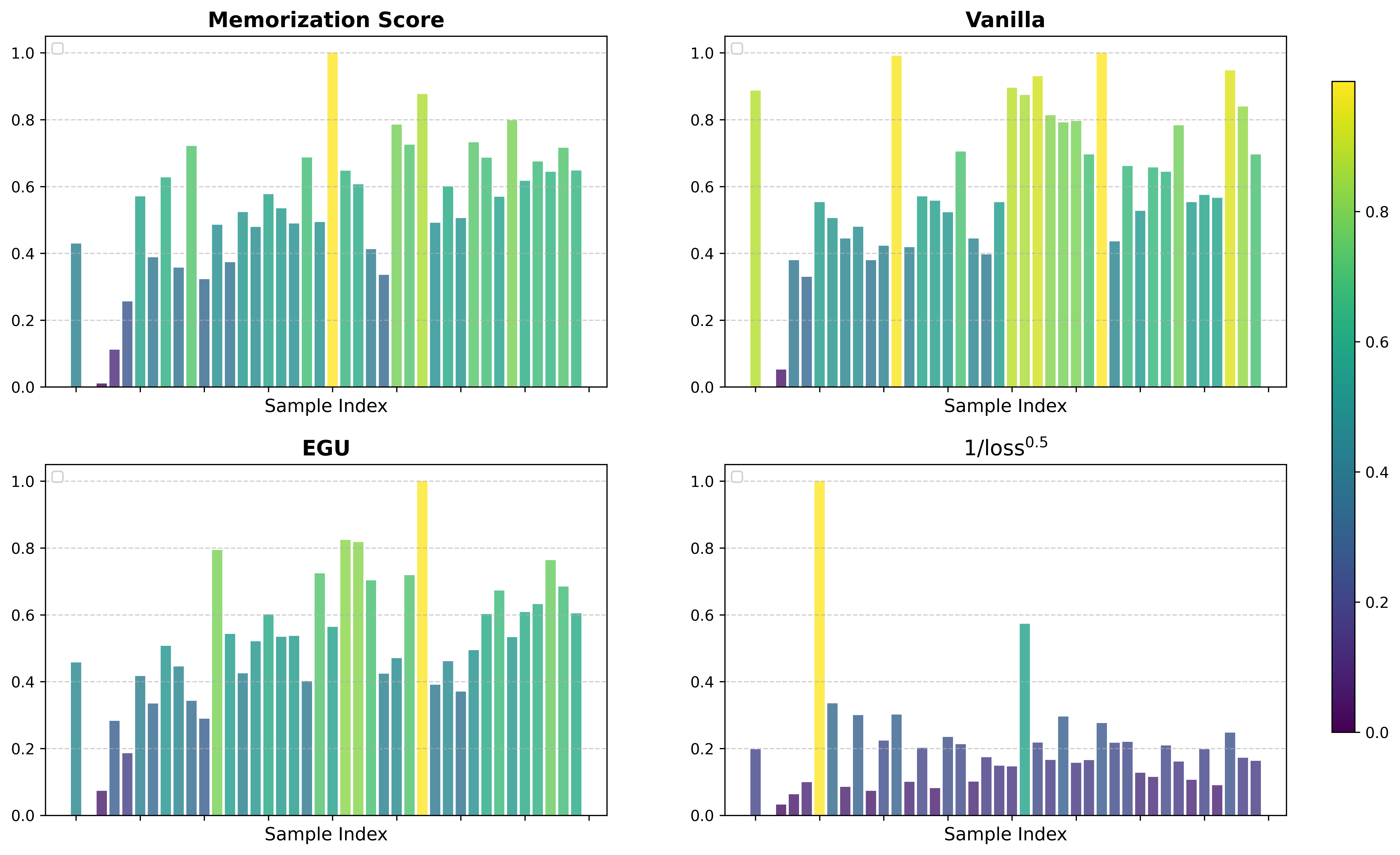}
    \caption{Normalized distributions of different unlearning strategies and memorization score. All distributions are min-max normalized for fair comparison. EGU most closely approximates the memorization score distribution.}
    \label{fig:me}
\end{figure}

\begin{figure}[t]
    \centering
    \includegraphics[width=\linewidth]{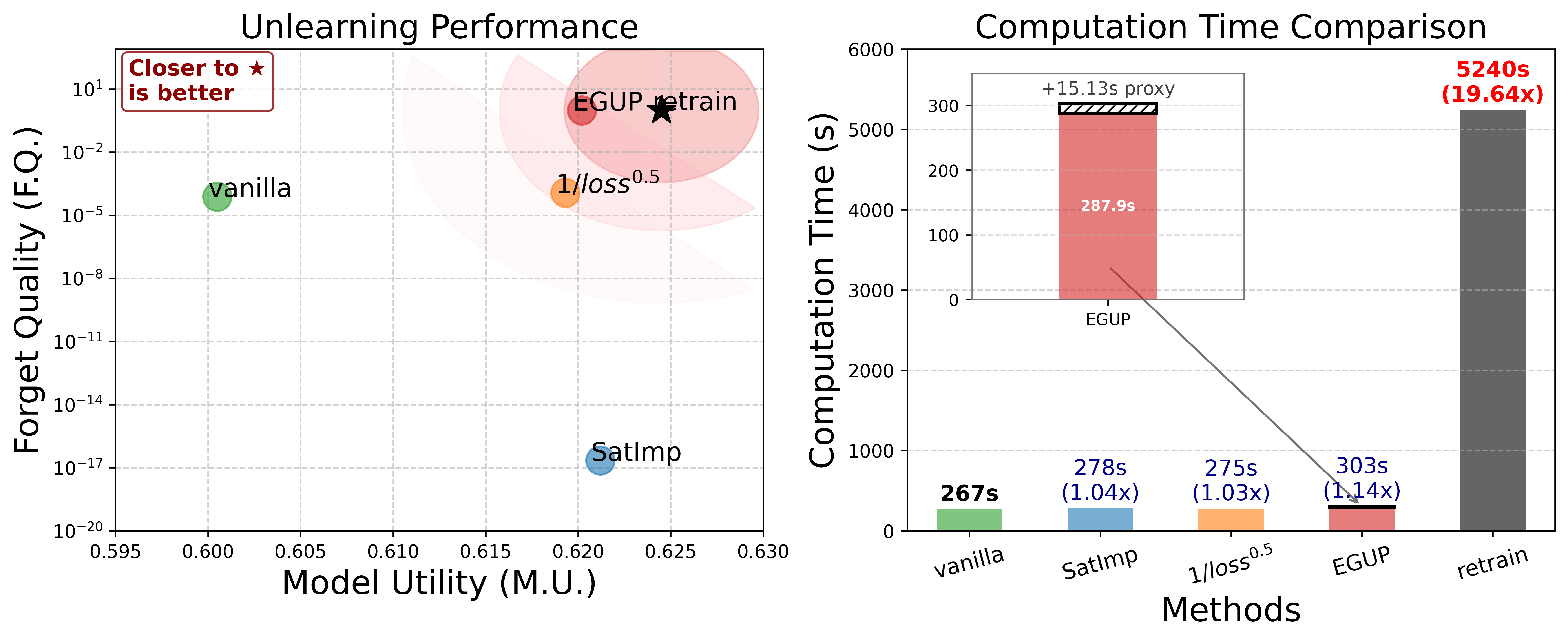}
    \caption{Comparison of different unlearning methods in terms of performance and computational cost under 5\% forgetting rate on LLaMA2-7B.}
    \label{fig:time}
\end{figure}

\begin{figure}[t]
    \centering
    \includegraphics[width=\linewidth]{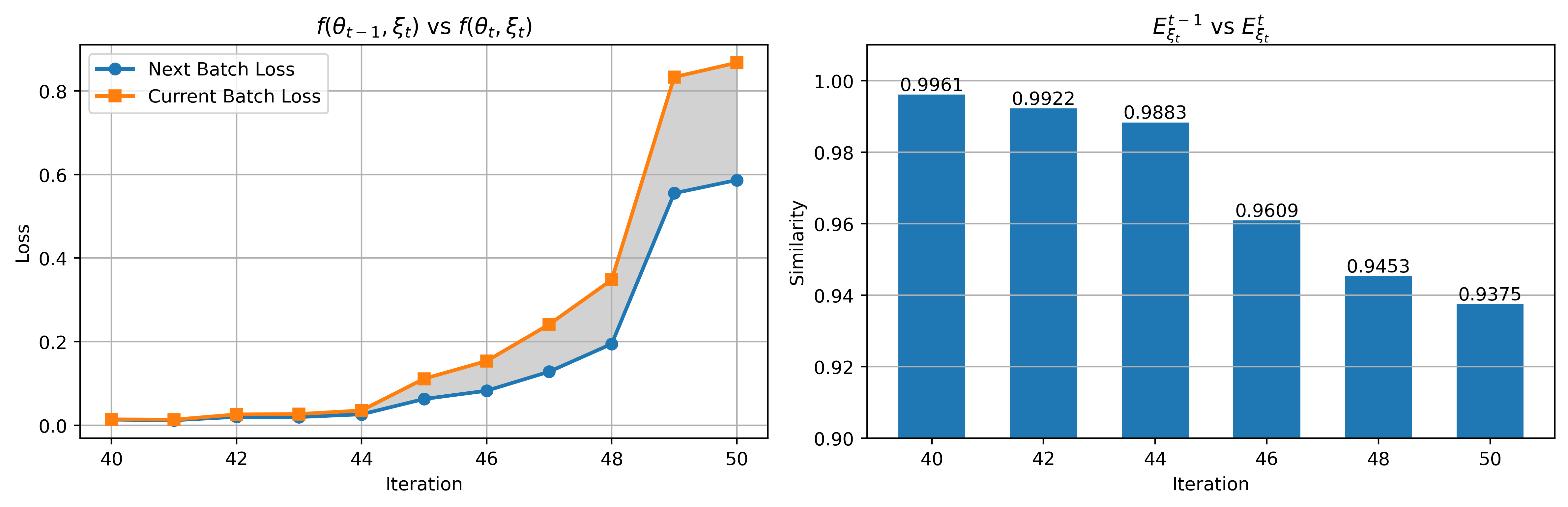}
    \caption{Cross-batch interference and representation drift during unlearning: as unlearning proceeds, loss on untouched samples rises while embedding similarity declines, indicating unintended over-unlearning.}
    \label{fig:inf}
\end{figure}

Neural networks do not memorize all samples equally, some examples are deeply memorized, while others are easily forgotten \cite{feldman2020neural, toneva2019empirical}. Consequently, effective unlearning requires adaptive control over the unlearning process across different dimensions. However, existing approaches often fail to achieve such fine-grained control, leading to inefficiency or over-unlearning. We summarize these challenges from three complementary perspectives: sample-level, iteration-level, and overall-level unlearning effort.

\partitle{Sample-level: determining how much to forget for each sample} 
In practice, we first observe that unlearning effort should varies substantially across samples: some examples are tightly entangled with retain data, while others can be safely removed without affecting model utility. As shown in Figure~\ref{fig:me}, samples with higher entanglement tend to exhibit stronger memorization behaviors, indicating that the model has overfitted to those specific instances, which suggests that entanglement can serve as a useful signal for estimating how much each sample should be forgotten. Specifically, the update induced by EGU exhibits a smaller Euclidean distance (1.0901 vs. 1.2892), suggesting that its update magnitude is also closer to that of retraining. Instead of explicitly measuring memorization, which typically requires retraining or influence-based analysis, we introduce an entanglement-guided loss reweighting mechanism that provides a lightweight and effective alternative. As shown in Figure~\ref{fig:time}, EGUP achieves better unlearning performance while maintaining low computational overhead. Specifically, EGUP takes 287.9s for the unlearning process, and the additional ICL-based proxy generation costs only 15.13s, which can be performed offline or amortized through batched prompting. This enables efficient and fine-grained control over sample-level unlearning while remaining substantially cheaper than full retraining.

\partitle{Iteration-level: adapting unlearning effort over time} 
As shown in Figure \ref{fig:inf}, updates applied to one forget batch inevitably propagate to others through the shared parameter space, leading to cross-batch interference. As unlearning progresses, we observe a continuous increase in the loss of untouched samples, together with a steady decline in embedding similarity. This indicates a global representation drift rather than a localized adjustment, meaning that unlearning effects are not confined to targeted samples. These dynamics show that LLM unlearning is inherently coupled across samples and optimization steps, where parameter shifts can propagate and amplify, increasing the risk of over-unlearning and destabilizing the representation space. Although such representation shifts are generally undesirable side effects of unlearning, they provide valuable signals that, if properly exploited, can be used to more effectively guide and regulate the unlearning process.

\begin{figure}[t]
    \centering
    \includegraphics[width=0.8\linewidth]{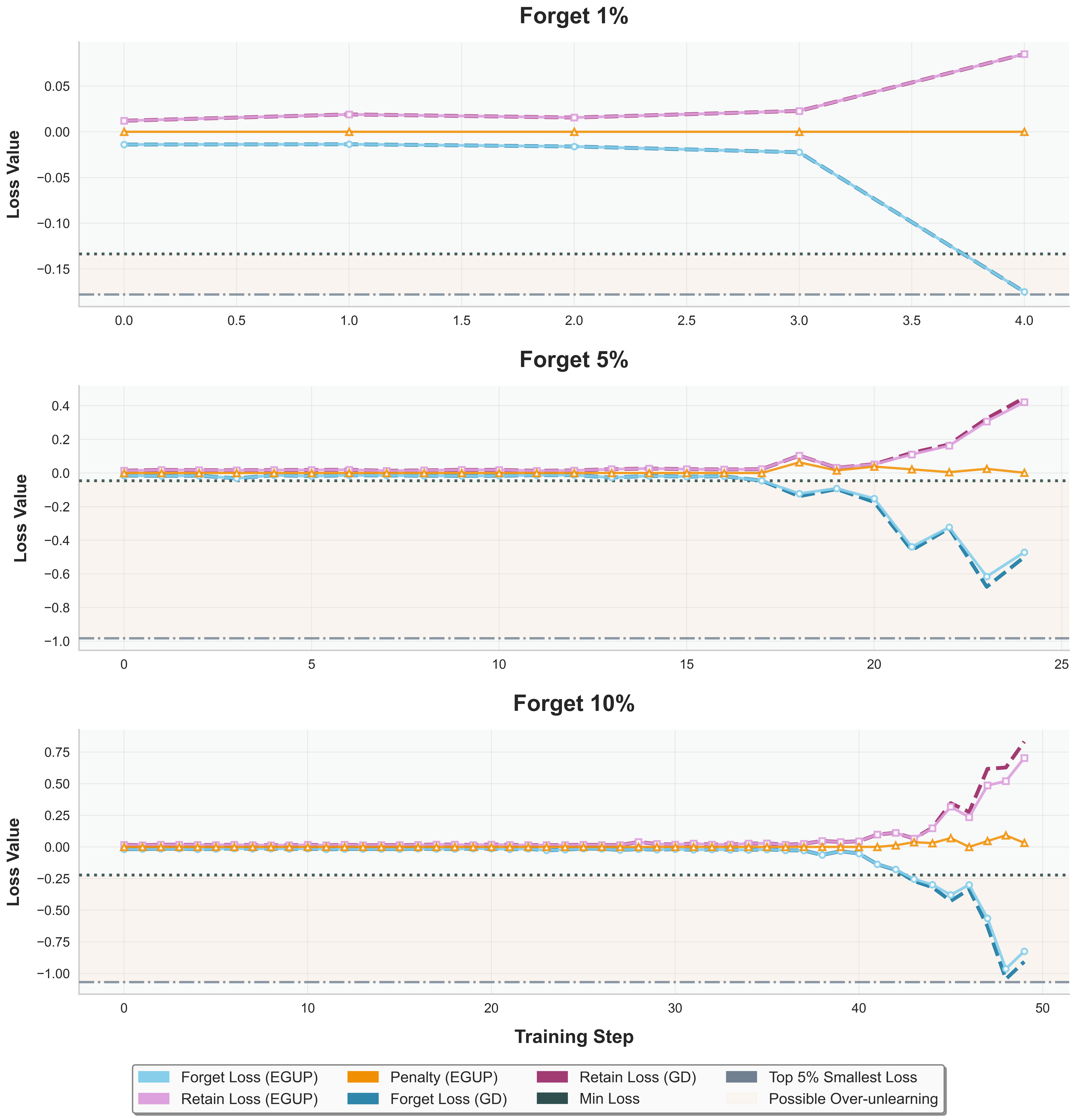}
    \caption{Forgetting dynamics at 1\%, 5\%, and 10\% comparing GD (dashed) with EGUP\_w/o (GD) (solid). Standard GD occasionally leads to excessive unlearning.}
    \label{fig:test}
\end{figure}

\begin{figure*}[tbp]
    \centering
    \includegraphics[width=0.9\linewidth]{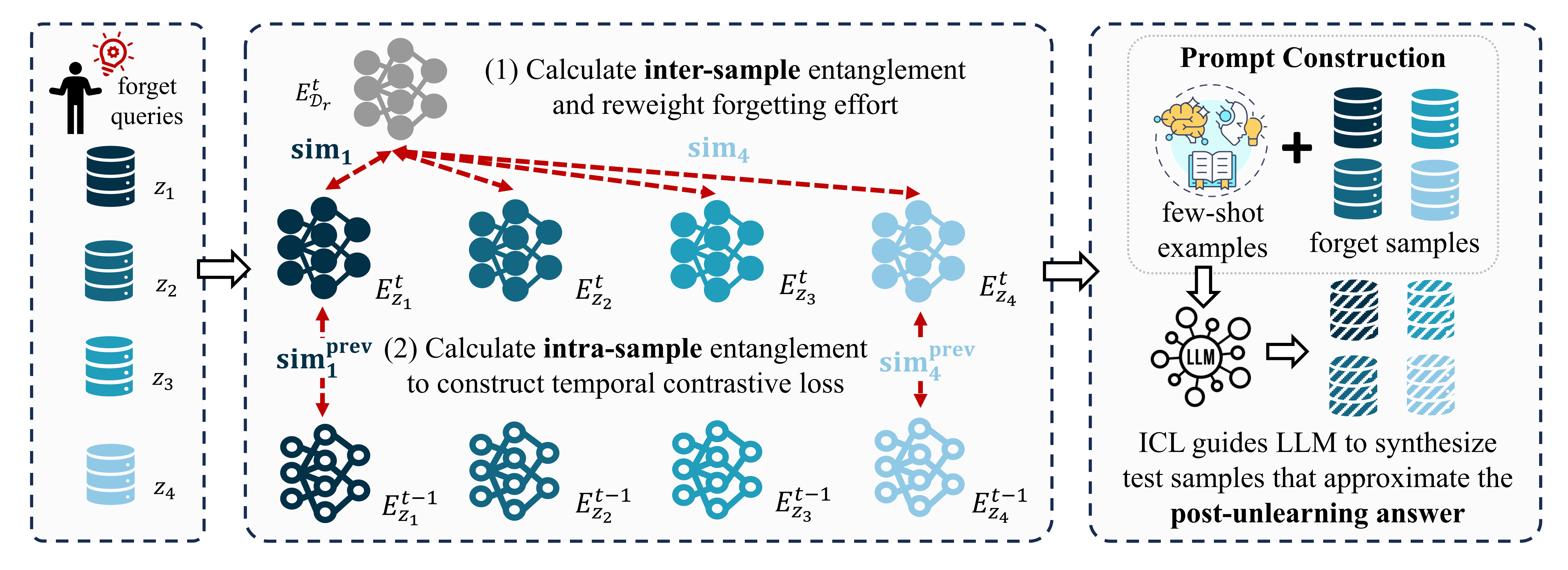}
    \caption{Overview of the EGUP framework, which integrates dual entanglement‑guided unlearning with a proxy constraint to steer and stabilize the unlearning process, ensuring effective forgetting while preserving essential knowledge.}
    \label{fig:workflow}
\end{figure*}

\partitle{Overall-level: determining when to stop unlearning} 
A key challenge in unlearning lies in the fact that the target degree of unlearning is inherently undefined. During the unlearning process, the model has no clear indication of whether the current level of unlearning is sufficient or already excessive. As shown in Figure~\ref{fig:test}, the forget loss keeps increasing during the unlearning process and eventually exceeds its corresponding loss under the retrained model, indicating that the model continues to push forget samples away even beyond what is necessary for effective unlearning. This uncontrolled behavior often leads to over-unlearning, where certain samples are excessively altered beyond what is necessary. Therefore, it becomes essential to design mechanisms that enable the model to adaptively recognize when to stop unlearning, achieving a balance between effective removal and distributional stability.

\section{Methodology}

In real-world scenarios, models do not memorize all samples equally, making it difficult to determine how much unlearning effort each sample requires. EGUP addresses this by introducing an \textbf{entanglement-guided perspective}, where the unlearning effort is adaptively determined by how strongly a forget sample is entangled with retained knowledge. Yet the model still lacks a reliable behavioral reference indicating when unlearning should stop. EGUP therefore further introduces an external behavioral constraint that \textbf{approximates the desired “knowledge-removed” outputs}, providing a soft stopping signal and stabilizing the unlearning process. The overall workflow of EGUP is illustrated in Figure \ref{fig:workflow}. 

\subsection{Entanglement-Guided Unlearning (EGU)}
In this paper, we propose \textbf{E}ntanglement-\textbf{G}uided \textbf{U}nlearning,  a scalable and principled framework that enables fine-grained control over the unlearning process. EGU leverages dual entanglement signals to guide unlearning: (1) \textbf{Inter-sample entanglement} quantifies the similarity between each forget embedding and the aggregated embedding of the retain dataset. Samples that are highly entangled with retained knowledge i.e., exhibiting higher embedding similarity are likely to be well-predicted by a retrained model, requiring minimal unlearning effort, whereas samples with lower entanglement are forgotten more aggressively. (2) \textbf{Intra-sample entanglement} captures the temporal evolution of each forget embedding across iterations. By continuously tracking how an embedding shifts across iterations even under cross-sample interference induced by updates on other forget samples within the shared parameter space—it adaptively modulates the unlearning process and promotes a smooth, stable trajectory of progressive unlearning.

Forget samples that are less entangled with retain dataset are typically semantically distant in the embedding space. These samples encode unique information with little overlap with the retain data, warranting stronger unlearning effort. EGU leverages this inter-sample entanglement signal to guide unlearning, approximating it via the cosine similarity between each forget sample and the aggregated embedding of the retain dataset, and adaptively modulating unlearning effort for more precise and effective removal of the target knowledge. In addition, intra-sample entanglement tracks the temporal evolution of each forget embedding, adaptively shaping the unlearning trajectory to ensure smooth and stable unlearning progresses. Together, inter- and intra-sample entanglement enable EGU to perform precise and stable unlearning. The overall framework is detailed in Algorithm \ref{alg1}. Notably, the framework is designed to be modular and flexible, allowing seamless integration into existing unlearning methods, including gradient ascent \cite{jang2022knowledge}, gradient difference \cite{yao2024large}, negative preference optimization (NPO) \cite{zhang2024negative}, and NPO+GD (NPO with gradient descent on the retain dataset).

\begin{algorithm}[t]
\caption{Entanglement-Guided Unlearning with Proxy Constraint}\label{alg1}
\begin{algorithmic}[1]
\STATE \textbf{Input:} Retain dataset $\mathcal{D}_r=\left\{z_j\right\}_{j=1}^{|\mathcal{D}_r|}=\left\{(x_j,y_j)\right\}_{j=1}^{|\mathcal{D}_r|}$, Forget dataset $\mathcal{D}_f=\left\{z_i\right\}_{i=1}^{|\mathcal{D}_f|}=\left\{(x_i,y_i)\right\}_{i=1}^{|\mathcal{D}_r|}$, model loss function $f(\theta)$, retain strength $\alpha$, contrastive strength $\mu$
\STATE \textbf{Output:} Updated model parameters $\theta$

\STATE $\theta_0=\theta^{o}$

\FOR{$t=0$ to $T-1$}
    \STATE $E_{\mathcal{D}_r}^{t} \gets \frac{1}{|\mathcal{D}_r|} \sum_{x_j \in\mathcal{D}_r} \text{AvgTokenEmbedding}(\theta_t,x_j)$

    \STATE Sample $\xi_t=\left\{z_i\right\}_{i=1}^{m}$ from $\mathcal{D}_f$
    \FOR{$\{x_i,y_i\} \in \xi_t$}
    \STATE $E_{x_i}^t \gets \text{AvgTokenEmbedding}(\theta_t,x_i)$
    \STATE $\mathrm{sim}_{i} \gets  \langle E_{x_i}^t, E_{\mathcal{D}_r}^{t} \rangle / \left(\| E_{x_i}^t \|_2  \cdot \| E_{\mathcal{D}_r}^{t} \|_2\right) $
    \STATE $\mathrm{sim}_{i}^{\text{prev}} \gets  \langle E_{x_i}^t, E_{x_i}^{t-1} \rangle / \left(\| E_{x_i}^t \|_2  \cdot \| E_{x_i}^{t-1} \|_2\right) $
    \STATE $w_i \gets \exp(-\mathrm{sim}_i)$
    \ENDFOR
    \STATE $w_i \gets \text{softmax}\big( k \cdot w_i \big)$ for all $i$

    \STATE $L_c(\theta_t)=- \log\frac{\exp({\mathrm{sim}_{i}})}{\exp({\mathrm{sim}_{i}})+\exp\left(\mathrm{sim}_{i}^{\text{prev}}\right
    )}$
    \STATE $L(\theta_t)=-\frac{1}{|\mathcal{D}_f|}\sum_{i=i}^{|\mathcal{D}_f|}w_i\cdot f(\theta_t,z_i) + \alpha\cdot\frac{1}{|\mathcal{D}_r|}\sum_{j=1}^{|\mathcal{D}_r|}f(\theta_t,z_j)+ \mu\cdot L_c(\theta_t)+\mathcal{P}(\xi_t,\mu)$
    \STATE $\theta_{t+1} \gets \theta_t - \eta \cdot \nabla L(\theta_t)$
\ENDFOR
\end{algorithmic}
\end{algorithm}

As shown in Algorithm~\ref{alg1}, the proposed EGUP framework performs unlearning through dual entanglement–aware optimization, guided by both inter- and intra-sample entanglement. At every iteration $t$ we first compute the sentence-level (average token) embedding for each sample in the retain dataset and derive a global retain embedding $E_{\mathcal{D}_r}^{t}$ by averaging over all retain samples. the model processes a mini-batch $\xi_t$ from the forget dataset. For each forget sample $x_i$, we compute its sentence-level embedding $E_{x_i}^t$ under the current parameters $\theta_t$, and measure two types of entanglement: (1) the \textbf{inter-sample} entanglement $\mathrm{sim}_i = \cos(E_{x_i}^t, E_{\mathcal{D}_r}^t)$, which quantifies the semantic entanglement between forget and retain representations; and (2) the \textbf{intra-sample} entanglement $\mathrm{sim}_i^{\text{prev}} = \cos(E_{x_i}^t, E_{x_i}^{t-1})$, which captures the parameter-dependent drift of forget embeddings across iterations. 

The overall unlearning objective integrates both inter- and intra-sample entanglement–guided optimization. The term $-\frac{1}{|\mathcal{D}_f|}\sum_i w_i \cdot f(\theta_t, x_i)$ enforces adaptive unlearning guided by inter-sample entanglement, reflecting the semantic correlation between forget and retain samples. Meanwhile, the $\mu\cdot\log\frac{\exp(\mathrm{sim}_i)}{\exp(\mathrm{sim}_i) + \exp(\mathrm{sim}_i^{\text{prev}})}$ regularizes intra-sample entanglement, adaptively shaping the unlearning trajectory according to the temporal progression of each sample’s representation. The model parameters are subsequently updated via gradient descent: $\theta_{t+1} = \theta_t - \eta \nabla_\theta L(\theta_t)$. Depending on the specific unlearning strategy, the loss term $f(\theta_t, x_i)$ can instantiate various unlearning objectives such as Gradient Ascent (GA) and Negative Preference Optimization (NPO), enabling flexible adaptation of EGUP to different unlearning paradigms.

To further clarify the motivation behind dual entanglement-guided optimization, note that the cosine similarity between forget and retain embeddings ranges from $-1$ to $1$. To ensure the coefficient remains positive and captures the negative correlation, we apply an exponential transformation $\exp(-cos)$, so that lower similarity (closer to -1) yields a larger coefficient. In this way, forget samples that are less similar to the retain dataset in the embedding space are assigned larger coefficients and thus receive stronger unlearning to better align with the behavior of the retrained model. Intuitively, when a forget sample closely resembles the retain dataset, the retrained model, which is trained without the forget sample, is likely to perform well on it, as the retain knowledge naturally generalizes to similar inputs. In the extreme case where a forget sample is almost identical to part of the retain dataset (e.g., a paraphrased sentence such as ``X is born in Y'' rewritten as ``The birthplace of X is Y''), the retrained model will naturally retain high performance on it. Therefore, samples with lower similarity to the retain data require stronger unlearning to remove residual knowledge and better approximate the behavior of the retrained model. Finally, the coefficients are normalized via softmax operation with a temperature parameter $k$ to flexibly control the unlearning effort and achieve adaptable unlearning behavior. 

Building upon this semantic weighting mechanism, the second term, $\mathrm{sim}_i^{\text{prev}} = \cos(E_{x_i}^t, E_{x_i}^{t-1})$, captures intra-sample entanglement, providing a temporal regularization that tracks and guides the evolution of each forget embedding throughout the unlearning process. In the early stages, when $\mathrm{sim}_i^{\text{prev}}$ tends to be large due to minimal parameter updates, the contrastive term yields a stronger gradient, accelerating representation divergence and amplifying the unlearning effect. As unlearning proceeds and embeddings progressively decorrelate, $\mathrm{sim}i^{\text{prev}}$ decreases, causing the contrastive penalty to diminish relative to the weighted unlearning loss. This gradual reduction naturally stabilizes optimization, preventing over-unlearning and maintaining representational coherence. By encouraging moderate divergence between $E_{x_i}^t$ and $E_{x_i}^{t-1}$, the temporal term thus promotes a smooth and consistent embedding trajectory that reflects progressive unlearning.

\subsection{EGU with Proxy Constraint (EGUP)}
While the entanglement-guided unlearning framework adaptively modulates the unlearning effort across samples and iterations based on inter-sample and intra-sample entanglement, it still faces a fundamental challenge in controlling the overall degree of unlearning. In practice, it is often difficult to \textbf{determine how much unlearning is sufficient}. Without a clear criterion, continued updates can excessively alter the model’s predictions on forget samples, pushing them away from responses to naturally unseen inputs. Such unnecessary divergence not only undermines generalization but also introduces distributional inconsistencies, increasing the risk of unlearning detection. This highlights the need for an appropriate stopping criterion to prevent such undesirable deviations during unlearning. However, gradient-ascent–based unlearning offers \textbf{no natural point of convergence}, making it inherently difficult to decide when the level of unlearning becomes sufficient for the intended objective. The optimal threshold can vary substantially across model architectures, training datasets, and the degree of entanglement between forget and retain data. A fixed threshold may be too conservative for some samples and too aggressive for others, leading to either under-unlearning or over-unlearning.

To this end, we propose EGUP, an extension of EGU that incorporates \textbf{P}roxy Constraint based on test inputs synthesized through in-context learning (ICL) with LLM. These proxy samples emulate how the model would naturally respond to forget samples if it has never been exposed to them. By applying a loss-based soft constraint on the model’s predictions over these proxy inputs during unlearning, EGUP provides a semantically grounded and adaptive boundary that not only eliminates the need for manual threshold tuning, but also effectively mitigates over-unlearning and preserves the model’s ability after unlearning.

Unlike traditional test data, which typically relies on real, unseen ground-truth labels, our approach introduces novel unlearning constraints by leveraging a language model to generate plausible reference answers, referred to as test answers. These test answers are generated via ICL, where the language model is prompted with carefully selected exemplars that contrast correct outputs with uninformed responses. This design enables the ICL-generated outputs to serve as high-quality surrogates that \textbf{approximate the behavior of the retrained model on unseen inputs}. As semantically aligned references, they offer an effective criterion for identifying over-unlearning in forget samples. An additional advantage of employing ICL-based proxy answers lies in their adaptability to \textbf{specific model behaviors} and \textbf{data distributions}. By carefully selecting in-context exemplars that mirror the target model’s response tendencies, both in terms of correct predictions and typical failure cases, the language model can simulate how the target model would behave when presented with similar inputs after unlearning. This enables the generated test data to more accurately reflect model-specific generalization under ignorance, resulting in a penalty signal better aligned.

During unlearning, we monitor the prediction loss of each forget sample with respect to its original label and compare it to the loss on the LLM-generated proxy answer. If the loss on the real answer surpasses that on the proxy baseline, we impose a penalty term that suppresses unlearning and prevents over-unlearning. This mechanism constrains the unlearning process to remain aligned with the behavior of the retrained model, serving as an \textbf{early warning signal} to prevent over-unlearning.

\begin{figure*}[htbp]
    \centering
    \includegraphics[width=0.9\linewidth]{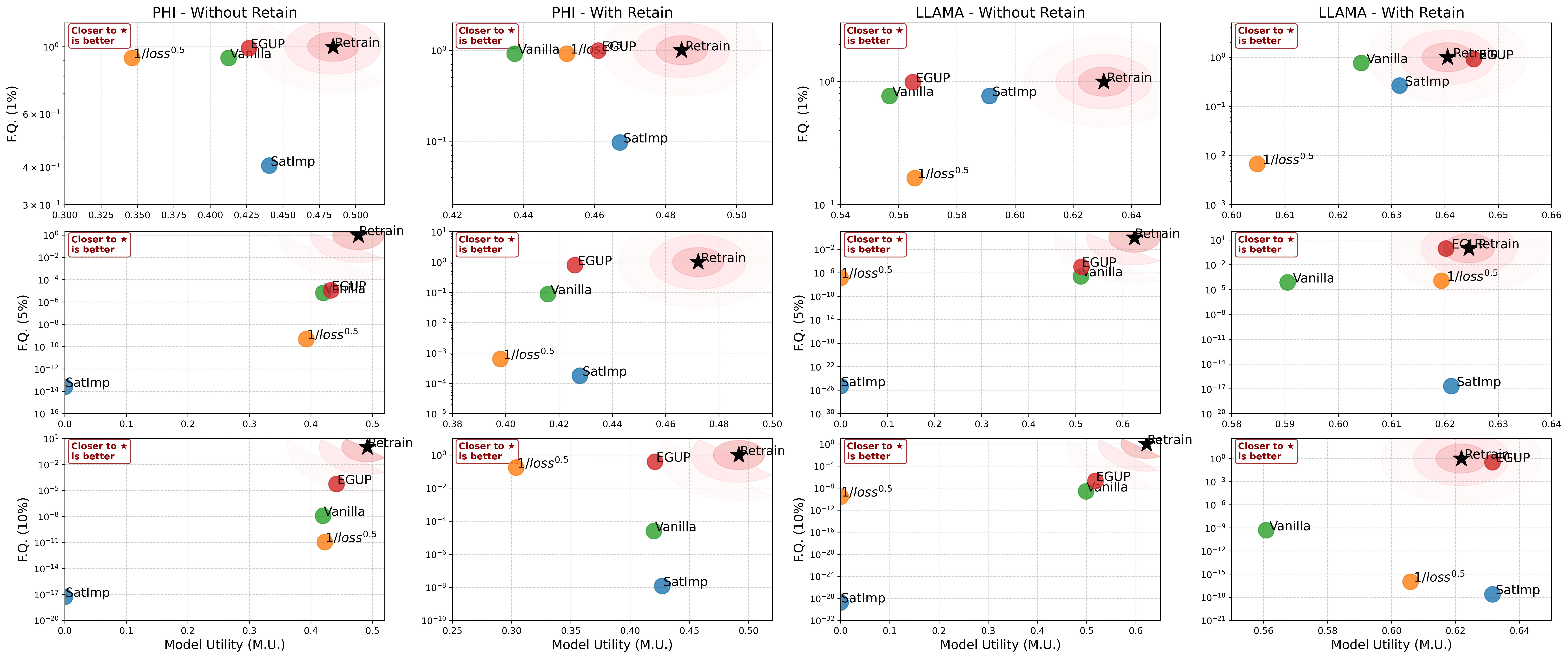}
    \caption{Comparison of unlearning performance on Phi‑1.5 and LLaMA2‑7B across Vanilla, SatImp \cite{yang2025exploring}, empirical‑loss‑based heuristics \cite{huang2024unified}, EGUP, and full retraining. EGUP shows performance closest to full retraining.}
    \label{fig:tofu_comparison}
\end{figure*}

\begin{algorithm}[tb]
\caption{Generating Proxy Answers and Computing over-unlearning Penalty $\mathcal{P}(\xi_t,\mu)$}
\label{alg3}
\begin{algorithmic}[1]
\STATE \textbf{Input:} Forget samples $\xi_t$, pretrained LLM (for ICL prompting), penalty weight $\mu$
\STATE \textbf{Step 1: Generate Proxy Answers}
\FOR{each forget data $z_i=\left\{x_i,y_i\right\}$ in $\xi_t$}
    \STATE Provide few-shot examples $\{(x_j, y_j, y_j')\}$ as prompt to the LLM
    \STATE Generate $y_i^{test}$: model’s expected answer without training on forget data
    \STATE Collect $(x_i, y_i^{test})$ as proxy sample $z_i^{test}$
\ENDFOR
\STATE Obtain proxy dataset $\mathcal{D}_{test} = \{(x_i, y_i^{test})\}$

\STATE \textbf{Step 2: Compute over-unlearning Penalty}
\STATE Initialize penalty $\mathcal{P} = 0$
\FOR{each forget sample $z_i=\left\{x_i, y_i\right\} \in\xi_t$}
    \STATE Compute forget loss: $f(\theta_t,z_i)$
    \STATE Compute proxy loss: $f(\theta_t,z_i^{test})$
    \STATE $\mathcal{P} \gets \mathcal{P} + \mu\cdot (f(\theta_t,z_i) - f(\theta_t,z_i^{test}))$ if $f(\theta_t,z_i) < f(\theta_t.z_i^{test})$
\ENDFOR
\STATE Return $\mathcal{P}$
\end{algorithmic}
\end{algorithm}

The detailed procedure is shown in Algorithm \ref{alg3}. In each iteration, given forget samples $\xi_t$, each sample $z_i = \{ x_i, y_i \}$ consists of a question $x_i$ and its ground-truth answer $y_i$. We first prepare few-shot exemplars $\{ (x_j, y_j, y_j') \}$ to prompt a pretrained large language model (LLM). Here, $y_j'$ denotes the model’s expected answer to $x_j$ without ever having seen the forget data. Conditioned on these exemplars, the LLM synthesizes a \emph{proxy answer} $y_i^{test}$ for the target question $x_i$. The resulting pair $z_i^{test} = \{ x_i, y_i^{test} \}$ serves as a test reference approximating the model’s natural prediction behavior under the absence of explicit memorization. All proxy samples are collected to form the test dataset $\mathcal{D}_{test}$. Next, for each forget sample $x_i$, we compute its current loss $f(\theta_t, z_i)$ and the loss of proxy data $f(\theta_t, z_i^{test})$. If the model’s loss on the forget data drops below the loss of proxy data $f(\theta_t, z_i) < f(\theta_t, z_i^{test})$, this indicates that the unlearning process may have overshot the target, deviating excessively from the model’s expected generalization. To softly penalize such over-unlearning, we accumulate a penalty $\mathcal{P}$ defined as: $\mathcal{P} \gets \mathcal{P} + \mu \cdot (f(\theta_t, z_i^{test}) - f(\theta_t, z_i)) \quad \text{if} \quad f(\theta_t, z_i) < f(\theta_t, z_i^{test})$, where $\mu$ is a hyper-parameter controlling the penalty’s strength. This penalty is then added to the overall unlearning objective.

In Algorithm \ref{alg3}, the examples used in ICL are formatted as question--answer (QA) pairs, following the convention established by the TOFU benchmark. Our method can also be naturally extended to other data modalities. For instance, narrative passages can be reformulated into QA pairs by framing key facts as questions and answers, while dialogue data can be adapted by converting turns into query–response pairs. While Algorithm \ref{alg3} presents the process as querying the LLM once per sample, in practice we adopt a more efficient strategy by \textbf{batching multiple forget samples into a single prompt}. This allows the LLM to generate corresponding proxy answers in parallel, substantially reducing the computational overhead and enhancing scalability to larger unlearning datasets. Below, we illustrate the prompt design used to obtain proxy answers through in-context learning (ICL). The prompt is structured into three components: System Instruction, which defines the model’s behavior; Input Description, which specifies the task context and information provided; and Output Requirement, which clarifies the expected form of the model’s response.

We present an example regarding fictitious author Basil Mahfouz Al-Kuwaiti, which includes the user query, the corresponding real (ground‑truth) answer, the pure answer produced by the model before acquiring the relevant knowledge, and the test answer generated via in‑context learning (ICL) using GPT‑4o. While the real answer identifies his genre as ``French Literature'', the pure answer suggests ``Alternate History'', and the ICL-based test answer produces ``Science Fiction''. As Alternate History is widely recognized as a \textbf{subgenre} of Science Fiction, the test answer exhibits \textbf{strong semantic alignment} with the pure answer. Both outputs deviate comparably from the real answer, suggesting that the ICL approach effectively approximates the behavioral characteristics of the retrained model, capturing its essential response tendencies.

\begin{table*}[t]
\centering
\caption{Results on TOFU benchmarks with different forgetting ratios (1\%, 5\%, and 10\%). Each method is evaluated using four metrics: F.Q., M.U., and R-L represent forget quality, model utility and ROUGE-L respectively.}\label{tab:tofu_results}
\resizebox{\linewidth}{!}{
\begin{tabular}{l|cc|cc|cc|cc|cc|cc}
\toprule
\multirow{2}{*}{\textbf{Method}} & \multicolumn{4}{c|}{\textbf{TOFU-1\%}} & \multicolumn{4}{c|}{\textbf{TOFU-5\%}} & \multicolumn{4}{c}{\textbf{TOFU-10\%}} \\
\cmidrule{2-13}
& \multicolumn{2}{c|}{\textbf{Forget Perf.}} & \multicolumn{2}{c|}{\textbf{Retain Perf.}} 
& \multicolumn{2}{c|}{\textbf{Forget Perf.}} & \multicolumn{2}{c|}{\textbf{Retain Perf.}} 
& \multicolumn{2}{c|}{\textbf{Forget Perf.}} & \multicolumn{2}{c}{\textbf{Retain Perf.}} \\
\cmidrule{2-13}
& F.Q. $\uparrow$ & R-L & M.U. $\uparrow$ & R-L 
& F.Q. $\uparrow$ & R-L & M.U. $\uparrow$ & R-L
& F.Q. $\uparrow$ & R-L & M.U. $\uparrow$ & R-L\\
\midrule
& \multicolumn{12}{c}{Phi-1.5} \\
\midrule

GA & 0.9900 & 0.2914 & 0.4151 & 0.4722 & 2.38e-6 & 0.3955 & 0.4075 & 0.3944 & 8.02e-5 & 0.3557 & 0.3878 & 0.3716 \\

NPO & 0.9188 & 0.2696 & 0.4126 & 0.4570 & 6.73e-6 & 0.4051 & 0.4203 & 0.4105 & 1.22e-8 & 0.3196 & 0.4197 & 0.3423 \\

GD & 0.9188 & 0.3193 & 0.4564 & 0.5456 & 1.39e-6 & 0.3480 & 0.4047 & 0.4160 & 1.40e-6 & 0.3738 & 0.4032 & 0.4033 \\

NPO+GD & 0.9188 & 0.2783 & 0.4375 & 0.5072 & 0.0878 & 0.3518 & 0.4158 & 0.3942 & 2.56e-5 & 0.4211 & 0.4200 & 0.4438 \\
\midrule
EGUP(GA) & \textbf{0.9999} & 0.3260 & 0.4282 & 0.4764 & 1.12e-5 & 0.4094 & 0.4261 & 0.4064 & 2.36e-4 & 0.3399 & 0.4106 & 0.3505 \\

EGUP(NPO) & 0.9900 & 0.2946 & 0.4266 & 0.4723 & 1.12e-5 & 0.3962 & 0.4327 & 0.4082 & 5.52e-5 & 0.3847 & \textbf{0.4417} & 0.3983 \\

EGUP(GD) & 0.9900 & 0.2884 & 0.4567 & 0.5186 & 1.12e-5 & 0.3621 & \textbf{0.4391} & 0.4351 &  5.52e-5 & 0.3356 &  0.4363 & 0.3925 \\ 
\rowcolor{cyan!10}
EGUP(NPO+GD) & 0.9900 & 0.3366 & \textbf{0.4610} & 0.5558 & \textbf{0.7934} & 0.3474 & 0.4259 & 0.4147 & \textbf{0.3958} & 0.3601 & 0.4209 & 0.4211 \\
\midrule
\rowcolor{magenta!10}
Retrain LLM & 1 & 0.4176 & 0.4845 & 0.6737 & 1 & 0.4202 & 0.4772 & 0.6740 & 1 & 0.4251 & 0.4917 & 0.7706 \\
\midrule

& \multicolumn{12}{c}{LLaMA2-7B} \\
\midrule

GA & 0.4046 & 0.4183 & 0.5590 & 0.7553 & 1.12e-5 & 0.4418 & 0.4921 & 0.4726 & 2.89e-11 & 0.0577 & 0 & 0.0756 \\

NPO & 0.7659 & 0.2905 & 0.5568 & 0.7187 & 2.61e-7 & 0.3885 & 0.5107 & 0.4625 & 2.55e-9 & 0.5009 & 0.4987 & 0.5082 \\

GD & 0.2657 & 0.2298 & 0.5255 & 0.4506 & 1.87e-9 & 0.0355 & 0.6106 & 0.7600 & 0.0017 & 0.2481 & 0.5281 & 0.3440 \\

NPO+GD & 0.7659 & 0.3793 & 0.6243 & 0.7737 & 7.54e-5 & 0.4559 & 0.5905 & 0.5681 & 5.02e-10 & 0.4490 & 0.5608 & 0.4815 \\
\midrule
EGUP(GA) & 0.9188 & 0.3719 & 0.5660 & 0.7762 & 1.12e-5 & 0.4808 & 0.5106 & 0.5112 & 1.16e-5 & 0.0018 & 0 & 0.0040 \\

EGUP(NPO) & \textbf{0.9900} & 0.3605 & 0.5648 & 0.7543 & 1.12e-5 & 0.4504 & 0.5119 & 0.4912 & 2.17e-7 & 0.4987 & 0.5178 & 0.5176 \\

EGUP(GD) & 0.7659 & 0.3391 & 0.5325 & 0.6814 & 2.96e-5 & 0.0000 & \textbf{0.6244} & 0.5694 & 0.0043 & 0.0761 & 0.6189 & 0.6659 \\
\rowcolor{cyan!10}
EGUP(NPO+GD) & 0.9188 & 0.2417 & \textbf{0.6454} & 0.8059 & \textbf{0.9647} & 0.3022 & 0.6202 & 0.7926 & \textbf{0.3417} & 0.2735 & \textbf{0.6315} & 0.7266 \\
\midrule
\rowcolor{magenta!10}
Retrain LLM & 1 & 0.3934 & 0.6305 & 0.9956 & 1 & 0.3930 & 0.6245 & 0.9810 & 1 & 0.3980 & 0.6218 & 0.9820 \\
\bottomrule
\end{tabular}
}
\end{table*}

\begin{tcolorbox}[
  colback=gray!4,
  colframe=black,
  title={\textbf{Prompt Template}},
  arc=2mm,
  boxrule=0.7pt,
  breakable
]

\textbf{System Instruction:}  
You are required to infer the \textit{pure\_answer}, meaning the answer a model would output without learning the relevant knowledge.

\medskip
\textbf{Input Description:}  
Each sample contains a question, the original answer, and a \textit{pure\_answer} produced by model that had never been exposed to the corresponding information. Your task is to analyze patterns of \textit{pure\_answer} and predict it for unseen questions.

\medskip
\textbf{Output Requirement:}  
Produce the most probable \textit{pure\_answer} rather than the true real-world answer.

\end{tcolorbox}

\section{Experiments}

In this section, we comprehensively evaluate EGUP under widely adopted benchmark settings for LLM unlearning. Specifically, we follow two representative scenarios: (i) removing knowledge related to a fictional author in the TOFU benchmark \cite{maini2024tofu}, and (ii) mitigating memorized or privacy-sensitive content in the MUSE benchmark \cite{shi2024muse}. Detailed experimental setups and evaluation metrics are provided in the appendix.

\subsection{Experiments Results}
We evaluate our unlearning method on the TOFU benchmark \cite{maini2024tofu}, which tests the removal of factual knowledge about 200 fictitious authors, each with 20 QA pairs. TOFU provides three forget sets $\mathcal{D}_f$ (1\%, 5\%, 10\% of authors), while the retain set $\mathcal{D}_r$ contains QA pairs of the remaining authors. Forget quality \cite{maini2024tofu} measures how closely the unlearned model matches one trained only on $\mathcal{D}_r$, whereas model utility reflects performance on held-out retain data, including information about fictional writers, real-world author profiles, and other world facts. Additionally, we report ROUGE-L scores for both forget and retain performance.

Experiments are conducted on fine-tuned Phi-1.5 \cite{li2023textbooks} and LLaMA2-chat-7B \cite{touvron2023llama}, both pretrained on all 200 authors. Table~\ref{tab:tofu_results} presents the experimental results on the TOFU benchmark, including comparisons across several unlearning methods: Gradient Ascent (GA), Gradient Difference (GD), Negative Preference Optimization (NPO), and their combination (NPO+GD), each reported in both vanilla form and when integrated with EGUP. The table highlights how EGUP consistently improves the vanilla baselines across different forgetting ratios. A broader comparison between EGUP and other state-of-the-art unlearning methods is illustrated in Figure~\ref{fig:tofu_comparison}.

\begin{figure*}[htbp]
    \centering
    \includegraphics[width=0.85\linewidth]{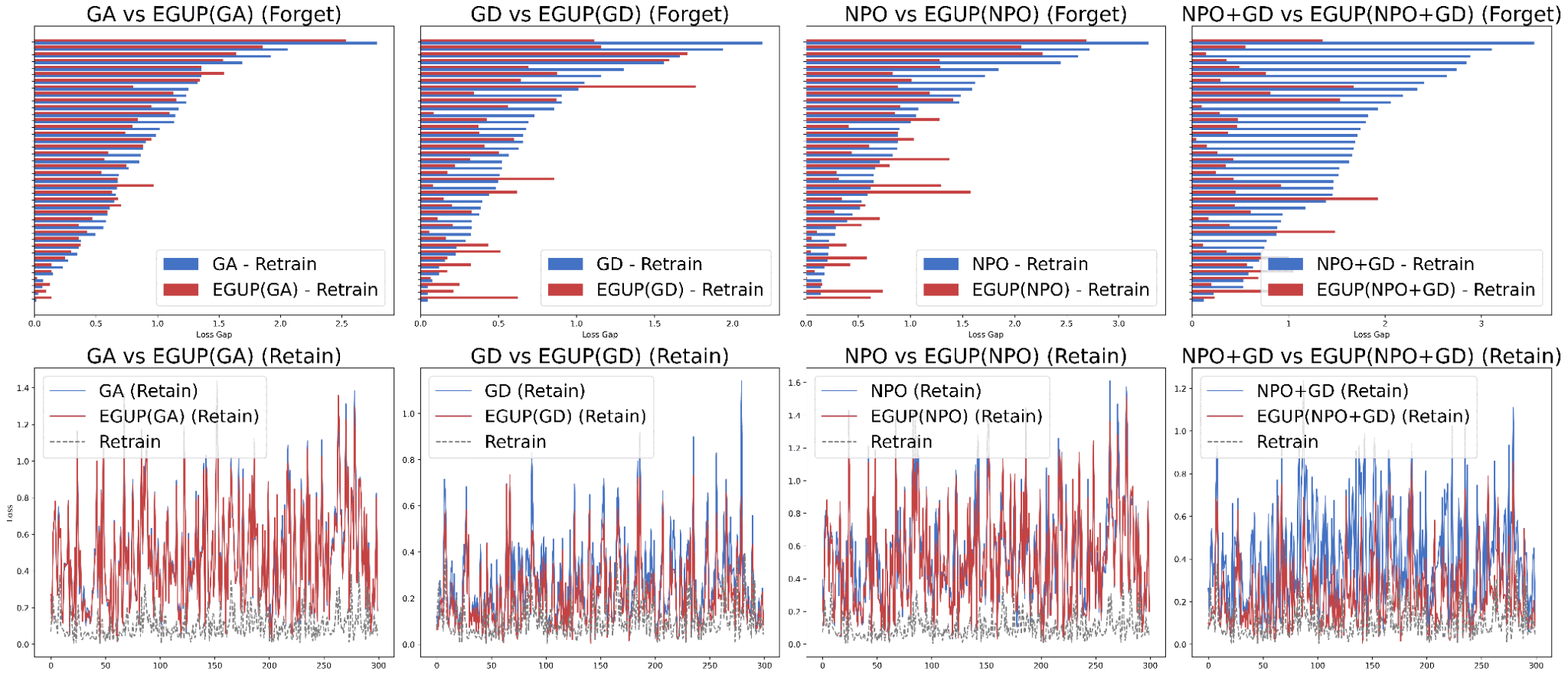}
    \caption{Loss difference on Forget 1\% between unlearned models and retrained model for each forget/retain sample using the Phi model. EGUP yield smaller loss gaps, indicating closer alignment with the retrained model.}
    \label{fig:loss_gap_phi}
\end{figure*}

\partitle{Consistent Performance Gains Across Metrics}
Our framework consistently outperforms the vanilla baselines across both Forget Quality and Model Utility on both Phi-1.5 and LLaMA2-7B in Table~\ref{tab:tofu_results}. On Phi-1.5, our framework consistently outperforms all vanilla baselines across both Forget Quality and Model Utility. Notably, for instance, under the 5\% setting, integrating NPO+GD into our framework significantly elevates the Forget Quality from \textbf{0.0878} to \textbf{0.7934}, while simultaneously improving Model Utility from \textbf{0.4158} to \textbf{0.4259}. Moreover, the ROUGE-L score on the retain set increases from \textbf{0.3942} to \textbf{0.4147}, demonstrating EGUP’s ability to maintain strong generalization while achieving effective unlearning. Similar trends are observed on the LLaMA2-7B model, confirming the generality and robustness of EGUP across architectures and scales. Under the 5\% unlearning setting, EGUP(NPO+GD) improves the Forget Quality from 7.54e-5 to \textbf{0.9647}, while also enhancing Model Utility from \textbf{0.5905} to \textbf{0.6202} and the Retain ROUGE-L from \textbf{0.5681} to \textbf{0.7926}. Importantly, consistent gains are observed across all other unlearning ratios as well. For example, under the 10\% setting, EGUP(NPO+GD) achieves a threefold increase in Forget Quality (0.3417 vs. 5.02e-10) while preserving substantially more useful knowledge, as evidenced by the higher Retain ROUGE-L (0.7266 vs. 0.4815) and Model Utility (0.6315 vs. 0.5608).

\partitle{Consistent Gains Across Optimizers}
Such consistent improvements are observed across all optimizers considered (e.g., GA, GD, NPO, and NPO+GD). Although GA performs less favorably than NPO, and GD underperforms compared to NPO+GD, the integration of all optimizers into our framework leads to noticeable gains. Notably, even though the trade-off between F.Q. and M.U. for GA under our framework remains inferior to the vanilla NPO baseline, it still benefits from our enhancements. We attribute this to the limited potential of the GA optimizer itself. In contrast, the combination of NPO+GD within our framework (denoted as EGUP(NPO+GD)) achieves forget quality that closely approximates that of the retrained model, demonstrating our framework’s strong capability in precise and efficient unlearning.

\partitle{EGUP framework requires only the average retain embedding yet outperforms methods that use the full retain dataset}
The EGUP framework enables even weaker optimizers to outperform stronger baselines by incorporating lightweight, entanglement guidance. A notable example is EGUP(GA), which builds upon GA and requires only the average embedding of the retain dataset. In contrast, GD depends on access to the full original retain dataset. Remarkably, EGUP(GA) consistently outperforms GD across multiple unlearning settings. Under the 5\% unlearning setting, EGUP(GA) achieves a higher model utility (0.4261 vs. 0.4047) than GD, while also slightly improving forget quality (1.12e-5 vs. 1.39e-6). This demonstrates that even coarse-grained retain signals, such as a single average embedding vector, can provide strong regularization to guide unlearning without explicit access to the full retain dataset or gradient-level control. The same trend holds across multiple forgetting ratios. For instance, under 10\% forgetting, EGUP(GA) boosts utility from 0.3878 to 0.4106, overtaking GD’s 0.4032 again, using only a minimal representation of the retain distribution.

\partitle{EGUP Consistently Outperforms Baselines Across Phi and LLaMA2-7B} 
Figure~\ref{fig:tofu_comparison} reports unlearning performance on both Phi and LLaMA2-7B across the TOFU benchmarks. Across the two model families, EGUP achieves a better trade-off between forget quality and model utility than alternative methods. The empirical-loss-based heuristic \cite{huang2024unified} is unstable: relative to the vanilla baseline it sometimes yields better unlearning performance but at other times performs worse, indicating high sensitivity to sample-level loss patterns and resulting variance across models and forgetting ratios. SatImp \cite{yang2025exploring} tends to preserve model utility well on both Phi and LLaMA, but its forget quality is notably weaker, suggesting incomplete removal of targeted knowledge.

\begin{figure*}[tbp]
    \centering
    \includegraphics[width=0.75\linewidth]{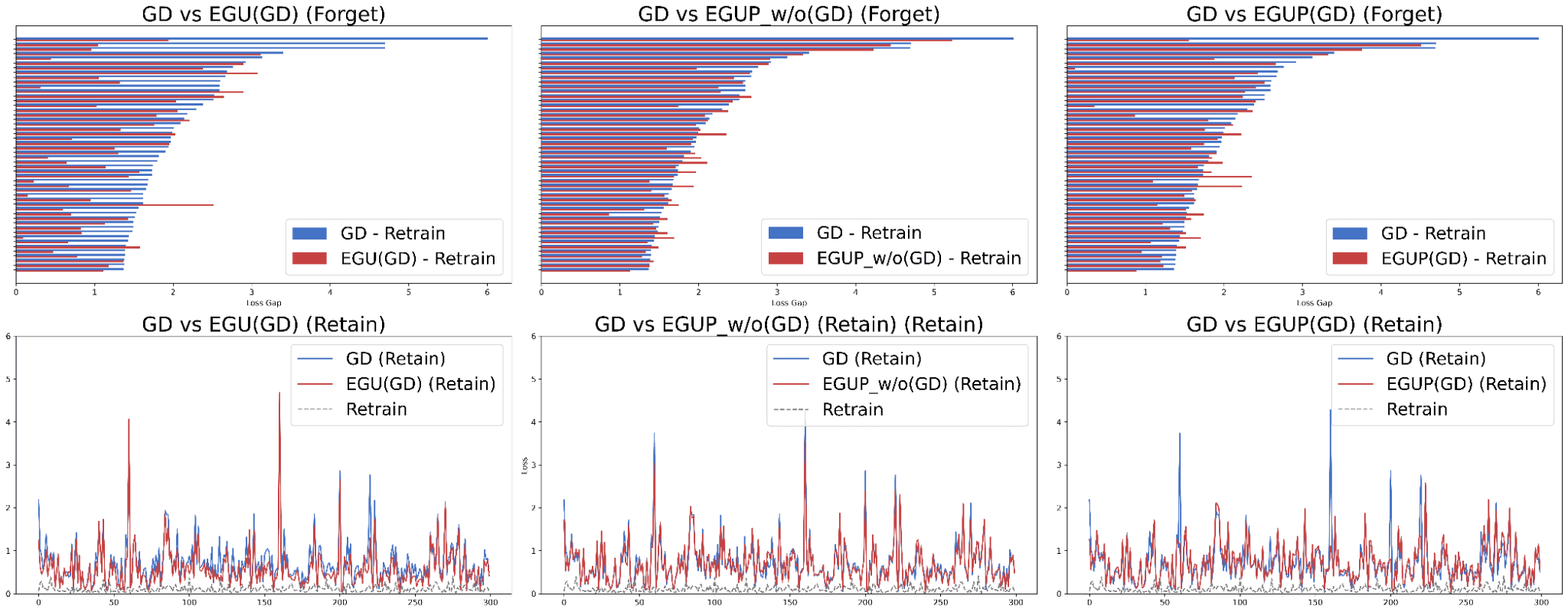}
    \caption{Forget and retain loss on Forget-05 are compared under different GD variants against retraining. EGU(GD) drops the proxy constraint and EGUP\_w/o(GD) drops the entanglement guidance. The results show that EGUP(GD) best balances unlearning and retention, closely approximating retraining.}
    \label{fig:ablation_loss}
\end{figure*}

We further illustrate the effectiveness of EGUP by visualizing, in Figure~\ref{fig:loss_gap_phi} the per-sample \textbf{absolute loss gap} between each unlearned model and the retrained model on both the forget and retain datasets. The top row shows results on the forget dataset, where we observe that models incorporating entanglement guided unlearning (EGU) consistently produce smaller loss gaps compared to their counterparts. EGU better approximate the behavior of retraining in removing memorized knowledge. The bottom row presents results on the retain dataset. Notably, methods built upon our entanglement-guided framework consistently yield loss patterns that more closely align with the retrained model, compared to their non-entanglement counterparts. This indicates that by assigning differentiated unlearning effort based on sample entanglement, our approach enhances unlearning precision while significantly mitigating undesired unlearning of the retain knowledge, thus achieving a more stable and utility-preserving unlearning process.

\subsection{Ablation Study}\label{sec:ablation}

\begin{figure}[t]
    \centering
    \includegraphics[width=\linewidth]{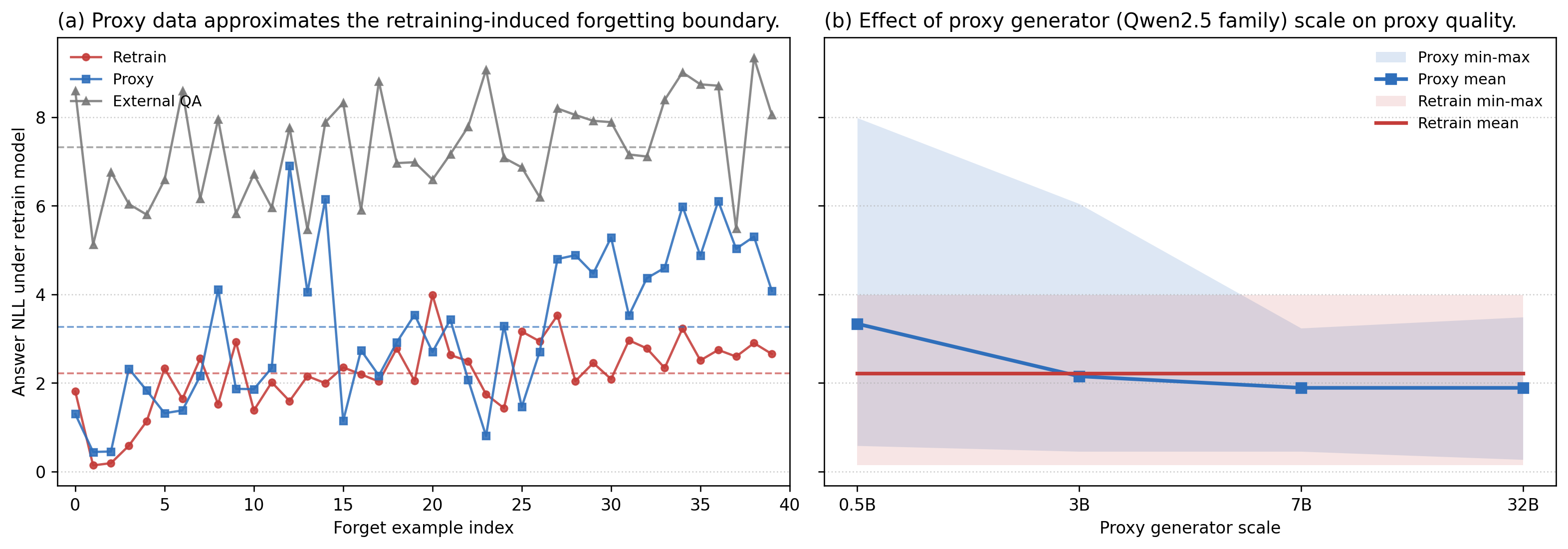}
    \caption{
Proxy data approximates the retraining-induced forgetting boundary.
(a) Under the retrained model, proxy examples yield losses close to original forget examples, while external QA examples remain outside the forgetting boundary.
(b) Larger proxy generators reduce the proxy loss range and better align the proxy mean with the original forget loss.
}
    \label{fig:proxy}
\end{figure}

\begin{table*}[t]
\centering
\caption{Evaluation of unlearning performance on the MUSE benchmark. EGUP effectively reduces both \textbf{C1} and \textbf{C3}, demonstrating improved control over memorization and privacy leakage.}
\resizebox{0.87\linewidth}{!}{
\label{tab:muse}
\renewcommand{\arraystretch}{1.15} 
\setlength{\tabcolsep}{4pt}        
\footnotesize                        
\begin{tabular}{
    >{\centering\arraybackslash}p{3.6cm}  
    >{\centering\arraybackslash}p{3.2cm}
    >{\centering\arraybackslash}p{3.2cm}
    >{\centering\arraybackslash}p{3.2cm}
    >{\centering\arraybackslash}p{3.2cm}
}
\toprule
\textbf{Method} & \textbf{C1. No Verbatim Mem.} & \textbf{C2. No Knowledge Mem.} & \textbf{C3. No Privacy Leak} & \textbf{C4. Utility Preserv.} \\
& VerbMem $\downarrow$ & KnowMem on $\mathcal{D}_{\text{forget}}$ $\downarrow$ & PrivLeak $\in [-5\%, 5\%]$ & KnowMem on $\mathcal{D}_{\text{retain}}$ $\uparrow$ \\
\midrule
\rowcolor{gray!20}
\multicolumn{5}{c}{\textbf{NEWS}} \\
\midrule
Target $f_{\text{target}}$ & 58.4 & 63.9 & -99.8 & 55.2 \\
\rowcolor{magenta!10}
Retrain $f_{\text{retrain}}$ & 20.8 & 33.1 & 0.0 & 55.0 \\
\midrule
GA & 0.0 & 0.0 & 24.7904 & 0.0 \\
\rowcolor{cyan!10}
EGUP(GA) & 15.5275~$\uparrow$ & 26.3561~$\uparrow$ & -21.7735~$\downarrow$ & 35.4112~$\uparrow$ \\
GD & 0.0256 & 24.9395 & 108.5289 & 22.3922 \\
\rowcolor{cyan!10}
EGUP(GD) & 0.0000~$\downarrow$ & 0.0000~$\downarrow$ & 29.2539~$\downarrow$ & 0.0000~$\downarrow$ \\
NPO & 51.0327 & 61.4428 & -99.8114 & 52.4327 \\
\rowcolor{cyan!10}
EGUP(NPO) & 42.5759~$\downarrow$ & 63.4412~$\uparrow$ & -99.7275~$\downarrow$ & 52.0777~$\downarrow$ \\
NPO+GD & 16.3903 & 37.1143 & -97.7578 & 37.4993 \\
\rowcolor{cyan!10}
EGUP(NPO+GD) & 5.6201~$\downarrow$ & 35.8123~$\downarrow$ & 83.2492~$\downarrow$ & 46.4921~$\uparrow$ \\
\midrule
\rowcolor{gray!20}
\multicolumn{5}{c}{\textbf{BOOKS}} \\
\midrule
Target $f_{\text{target}}$ & 99.8 & 59.4 & -57.5 & 66.9 \\
\rowcolor{magenta!10}
Retrain $f_{\text{retrain}}$ & 14.3 & 28.9 & 0.0 & 74.5 \\
\midrule
GA & 89.2175 & 37.7380 & -58.5759 & 56.7514 \\
\rowcolor{cyan!10}
EGUP(GA) & 84.3180~$\downarrow$ & 35.1689~$\downarrow$ & -54.8234~$\downarrow$ & 55.8213~$\downarrow$ \\
GD & 15.5676 & 11.6565 & 20.4317 & 19.0662 \\
\rowcolor{cyan!10}
EGUP(GD) & 0.0000~$\downarrow$ & 0.0000~$\downarrow$ & -24.8842~$\uparrow$ & 13.0053~$\downarrow$ \\
NPO & 16.1017 & 8.5882 & 31.0981 & 13.7530 \\
\rowcolor{cyan!10}
EGUP(NPO) & 13.9234~$\downarrow$ & 12.8425~$\uparrow$ & -29.0881~$\downarrow$ & 16.7468~$\uparrow$ \\
NPO+GD & 26.6586 & 26.4127 & -53.4026 & 43.8690 \\
\rowcolor{cyan!10}
EGUP(NPO+GD) & 7.8129~$\downarrow$ & 17.7289~$\downarrow$ & -35.5122~$\downarrow$ & 38.8175~$\downarrow$ \\
\bottomrule
\end{tabular}}
\end{table*}

To better understand the contribution of each component in EGUP, we conduct ablation studies by isolating the two key modules of EGUP. In Section~\ref{sec:loss_reweighting_comparison}, we evaluate the impact of entanglement-guided unlearning by disabling the proxy constraint. In Section~\ref{sec:icl}, we assess the effectiveness of the proxy constraint by removing EGU, thereby highlighting its role in regularizing the unlearning process and preventing over-unlearning.

\subsubsection{Entanglement-Guided Unlearning Framework}\label{sec:loss_reweighting_comparison}

To further evaluate the effect of entanglement-guided loss reweighting in our unlearning framework, we conduct an ablation study on Forget-05, using NPO as the base unlearning method. Figure~\ref{fig:ablation_loss} compares the losses on the forget and retain dataset against the retrained model. It also illustrates how each component, including entanglement-guided loss reweighting, proxy constraint, and their combination, impact the trade-off between unlearning efficacy and model utility.

In Figure~\ref{fig:ablation_loss}, each column presents the forget and retain losses under a specific variant of our method, evaluated against the retrained model's loss as a reference. The first column shows the effect of applying entanglement-guided unlearning (EGU) alone. This mechanism amplifies the influence of highly entangled tokens during unlearning, thereby improving the unlearning performance, evident from the forget loss being closer to that of the retrained model. However, this comes at the cost of increased retain loss, indicating potential over-unlearning or unintended interference with useful knowledge. The second column isolates the impact of proxy regularization based on in-context learning (ICL). Although it contributes little to reducing the forget loss, it substantially improves the retain loss, demonstrating its effectiveness in constraining over-unlearning and preserving model utility. Finally, the third column shows the result of EGUP. This combination achieves the best balance, with both forget and retain losses approaching those of the retrained model. The figure thus highlights the complementary strengths of the two modules and underscores that their integration is critical to achieve precise and utility-preserving unlearning.

\subsubsection{ICL-based Proxy Constraint}\label{sec:icl}

Figure~\ref{fig:proxy} evaluates whether the generated proxy data can approximate the forgetting boundary induced by retraining. 
Figure \ref{fig:test} compares the unlearning dynamics under three forgetting rates (1\%, 5\%, 10\%) between standard gradient difference (dashed) and EGUP\_w/o(GD) (solid), across forget loss, retain loss, and penalty loss. We summarize our findings through the following key observations.

\partitle{Proxy data approximates the retraining-induced forgetting boundary} 
As shown in Figure~\ref{fig:proxy}(a), under the retrained model, the answer NLLs of proxy examples closely track the loss pattern induced by retraining. In contrast, external QA examples incur substantially larger losses and remain well outside the forgetting region. This indicates that the generated proxy data captures the loss behavior of the retrained model, rather than merely behaving like generic external QA data (SQuAD).

\partitle{Larger proxy generators improve approximation quality}
Figure~\ref{fig:proxy}(b) further shows that the scale of the proxy generator directly affects proxy quality. As the generator scale increases, the proxy mean gradually aligns with the retraining forget mean, while the proxy min--max interval also shrinks and increasingly overlaps with the corresponding loss range under retraining. These results suggest that larger proxy generators produce proxy data whose loss distribution more closely matches the retraining reference distribution, thereby providing a more reliable surrogate constraint for approximating retraining.

\partitle{Stability in the early unlearning stage} 
At the initial training steps across all forget rates, both the forget loss and retain loss remain consistently low. This observation indicates that the unlearning process is well-behaved in its early phase. Meanwhile, the model maintains strong performance on the retain dataset, thus avoiding premature over-unlearning.

\partitle{No penalty activation under mild unlearning} In the 1\% forgetting case, no signs of over-unlearning are observed throughout the entire training trajectory. The penalty loss remains zero across all training steps, demonstrating that the test-informed penalty mechanism stays inactive when unnecessary and does not interfere with mild unlearning tasks.

\partitle{Proxy-based penalty anticipates over-unlearning} For the 5\% forgetting case, a sharp increase in retain loss is observed around step 18 for the GD baseline (dashed blue), signaling potential over-unlearning. Interestingly, the penalty loss (green) in EGUP\_w/o (GD) also rises concurrently, serving as an early warning signal. Notably, this penalty is not derived from the retain loss; instead, it is computed from test-time ICL proxy data and incorporated into the forget objective, enabling EGUP to proactively anticipate and mitigate over-unlearning before severe retain degradation occurs.

\partitle{ICL-informed penalty mitigates over-unlearning} In the later training stages, especially at higher forget rates (e.g., 5\% and 10\%), EGUP\_w/o (GD) (solid lines) maintains lower retain loss compared to the GD baseline. This confirms that integrating ICL proxy data into the loss formulation effectively regularizes the unlearning process and prevents excessive degradation of the retain knowledge, validating the effectiveness of our proxy-based penalty design.

\partitle{Limited proxy-generation overhead}
Figure~\ref{fig:time} reports the wall-clock computation time of different methods. Vanilla unlearning takes 267s, while EGUP takes 288s in total. The additional ICL-based proxy generation costs only 15.13s, accounting for a small fraction of the overall EGUP runtime. In contrast, full retraining requires 5240s, which is 18.19$\times$ slower than EGUP. These results show that EGUP introduces only limited overhead over vanilla unlearning, while remaining substantially more efficient than full retraining. The proxy-generation cost scales approximately linearly with the number of forget examples or proxy queries, since each proxy answer can be generated independently and can be amortized through batched prompting.

\subsection{Experiments on MUSE}\label{exp:muse}
We further extend our analysis to the MUSE benchmark \cite{shi2024muse}, incorporating additional evaluation metrics to provide a more comprehensive evaluation. MUSE considers two representative types of textual data that may frequently involve unlearning requests: the News and Books subsets. The News subset comprises BBC news articles published after August 2023, which are partitioned into disjoint forget, retain, and holdout sets. The Books subset includes the Harry Potter book series. In this benchmark, both LLaMA-2-7B and ICLM-7B serve as pre-trained base models. In this work, we concentrate on Verbatim Memorization, Knowledge Memorization, Privacy Leakage, and Utility Preservation to evaluate unlearning effectiveness and safety, following the metrics defined in MUSE. 

As shown in Table~\ref{tab:muse}, EGUP consistently improves unlearning effectiveness across both the NEWS and BOOKS domains. Specifically, it significantly reduces the \textbf{C1} and \textbf{C3} metrics, which measures verbatim memorization and privacy leakage. This demonstrates its capacity to mitigate both surface-level and sensitive-data retention. These improvements highlight EGUP’s advantage in achieving stronger deletion guarantees compared to baseline methods. However, these gains often come with reduced knowledge memorization on the retain dataset (\textbf{C4}). Additionally, we observe that EGUP’s performance varies across data domains, with BOOKS posing greater challenges due to longer sequences and richer semantics. This highlights the need for more fine-grained unlearning strategies. In this regard, we identify token-level reweighting as a promising future direction. By leveraging semantic similarity signals at finer granularity, such approaches could allow for more targeted unlearning at token level while better preserving unrelated capabilities, particularly in semantically dense domains.

\section{Conclusion}

In summary, this work presents EGUP, an entanglement-guided unlearning framework that enables more precise and controllable unlearning for LLMs. By jointly modeling inter-sample entanglement across samples and intra-sample entanglement across iterations, EGUP captures how forget samples are coupled with retain knowledge and how their representations evolve during unlearning. This dual entanglement guidance allows EGUP to adaptively modulate unlearning effort and mitigate over-unlearning. Beyond entanglement guidance, EGUP further addresses the stopping challenge by approximating retrained model behavior as a principled reference to regulate unlearning. Since obtaining a retrained model is often impractical due to its prohibitive cost, we instead leverage open-source LLMs as proxy estimators and employ in-context learning to generate proxy outputs that closely approximate retrain behavior for each forget sample. Overall, EGUP provides a modular and extensible perspective on entanglement-driven unlearning and shows great improvements over existing methods. Looking ahead, extending entanglement guidance to the token level and designing more accurate or self-adaptive proxy constraints represent promising directions for achieving finer-grained and more principled unlearning in the future.

\section*{Acknowledgment}
We would like to thank the anonymous reviewers for their helpful comments. This research is supported in part by the “Pioneer” and “Leading Goose” R\&D Program of Zhejiang (Grant No. 2024C01169), the Kunpeng–Ascend Science and Education Innovation Excellence/Incubation Center, the National Natural Science Foundation of China (Grant No. 62441238), and the National Natural Science Foundation of China under Grant U2441240 (“Ye Qisun” Science Foundation).
\bibliographystyle{plain} 
\bibliography{bib}

@inproceedings{feldman2020neural,
  title={What neural networks memorize and why: Discovering the long tail via influence estimation},
  author={Feldman, Vitaly},
  booktitle={Advances in Neural Information Processing Systems (NeurIPS)},
  volume={33},
  pages={2881--2891},
  year={2020}
}

@inproceedings{toneva2019empirical,
  title={An empirical study of example forgetting during deep neural network learning},
  author={Toneva, M. and Sordoni, A. and Tachet des Combes, R. and Trischler, A. and Bengio, Y. and Gordon, G. J.},
  booktitle={International Conference on Learning Representations (ICLR)},
  year={2019}
}

@article{huang2024unified,
  title={Unified Gradient-Based Machine Unlearning with Remain Geometry Enhancement},
  author={Huang, Zhehao and Cheng, Xinwen and Zheng, JingHao and Wang, Haoran and He, Zhengbao and Li, Tao and Huang, Xiaolin},
  journal={arXiv preprint arXiv:2409.19732},
  year={2024}
}

@article{eldan2023s,
  title={Who’s harry potter? approximate unlearning for LLMs},
  author={Eldan, Ronen and Russinovich, Mark},
  year={2023}
}

@article{maini2024tofu,
  title={Tofu: A task of fictitious unlearning for llms},
  author={Maini, Pratyush and Feng, Zhili and Schwarzschild, Avi and Lipton, Zachary C and Kolter, J Zico},
  journal={arXiv preprint arXiv:2401.06121},
  year={2024}
}

@inproceedings{feldman2020does,
  title={Does learning require memorization? a short tale about a long tail},
  author={Feldman, Vitaly},
  booktitle={Proceedings of the 52nd annual ACM SIGACT symposium on theory of computing},
  pages={954--959},
  year={2020}
}

@article{wei2024evaluating,
  title={Evaluating copyright takedown methods for language models},
  author={Wei, Boyi and Shi, Weijia and Huang, Yangsibo and Smith, Noah A and Zhang, Chiyuan and Zettlemoyer, Luke and Li, Kai and Henderson, Peter},
  journal={Advances in Neural Information Processing Systems},
  volume={37},
  pages={139114--139150},
  year={2024}
}

@article{yao2024survey,
  title={A survey on large language model (llm) security and privacy: The good, the bad, and the ugly},
  author={Yao, Yifan and Duan, Jinhao and Xu, Kaidi and Cai, Yuanfang and Sun, Zhibo and Zhang, Yue},
  journal={High-Confidence Computing},
  volume={4},
  number={2},
  pages={100211},
  year={2024},
  publisher={Elsevier}
}

@article{liu2023jailbreaking,
  title={Jailbreaking chatgpt via prompt engineering: An empirical study},
  author={Liu, Yi and Deng, Gelei and Xu, Zhengzi and Li, Yuekang and Zheng, Yaowen and Zhang, Ying and Zhao, Lida and Zhang, Tianwei and Wang, Kailong and Liu, Yang},
  journal={arXiv preprint arXiv:2305.13860},
  year={2023}
}

@article{zhao2024makes,
  title={What makes unlearning hard and what to do about it},
  author={Zhao, Kairan and Kurmanji, Meghdad and B{\u{a}}rbulescu, George-Octavian and Triantafillou, Eleni and Triantafillou, Peter},
  journal={Advances in Neural Information Processing Systems},
  volume={37},
  pages={12293--12333},
  year={2024}
}

@inproceedings{bourtoule2021machine,
  title={Machine unlearning},
  author={Bourtoule, Lucas and Chandrasekaran, Varun and Choquette-Choo, Christopher A and Jia, Hengrui and Travers, Adelin and Zhang, Baiwu and Lie, David and Papernot, Nicolas},
  booktitle={2021 IEEE symposium on security and privacy (SP)},
  pages={141--159},
  year={2021},
  organization={IEEE}
}

@inproceedings{cao2015towards,
  title={Towards making systems forget with machine unlearning},
  author={Cao, Yinzhi and Yang, Junfeng},
  booktitle={2015 IEEE symposium on security and privacy},
  pages={463--480},
  year={2015},
  organization={IEEE}
}

@article{zhang2024negative,
  title={Negative preference optimization: From catastrophic collapse to effective unlearning},
  author={Zhang, Ruiqi and Lin, Licong and Bai, Yu and Mei, Song},
  journal={arXiv preprint arXiv:2404.05868},
  year={2024}
}

@article{jang2022knowledge,
  title={Knowledge unlearning for mitigating privacy risks in language models},
  author={Jang, Joel and Yoon, Dongkeun and Yang, Sohee and Cha, Sungmin and Lee, Moontae and Logeswaran, Lajanugen and Seo, Minjoon},
  journal={arXiv preprint arXiv:2210.01504},
  year={2022}
}

@article{yao2024large,
  title={Large language model unlearning},
  author={Yao, Yuanshun and Xu, Xiaojun and Liu, Yang},
  journal={Advances in Neural Information Processing Systems},
  volume={37},
  pages={105425--105475},
  year={2024}
}

@article{yang2025exploring,
  title={Exploring Criteria of Loss Reweighting to Enhance LLM Unlearning},
  author={Yang, Puning and Wang, Qizhou and Huang, Zhuo and Liu, Tongliang and Zhang, Chengqi and Han, Bo},
  journal={arXiv preprint arXiv:2505.11953},
  year={2025}
}

@article{li2023textbooks,
  title={Textbooks are all you need ii: phi-1.5 technical report},
  author={Li, Yuanzhi and Bubeck, S{\'e}bastien and Eldan, Ronen and Del Giorno, Allie and Gunasekar, Suriya and Lee, Yin Tat},
  journal={arXiv preprint arXiv:2309.05463},
  year={2023}
}

@article{touvron2023llama,
  title={Llama 2: Open foundation and fine-tuned chat models},
  author={Touvron, Hugo and Martin, Louis and Stone, Kevin and Albert, Peter and Almahairi, Amjad and Babaei, Yasmine and Bashlykov, Nikolay and Batra, Soumya and Bhargava, Prajjwal and Bhosale, Shruti and others},
  journal={arXiv preprint arXiv:2307.09288},
  year={2023}
}

@article{shi2024muse,
  title={Muse: Machine unlearning six-way evaluation for language models},
  author={Shi, Weijia and Lee, Jaechan and Huang, Yangsibo and Malladi, Sadhika and Zhao, Jieyu and Holtzman, Ari and Liu, Daogao and Zettlemoyer, Luke and Smith, Noah A and Zhang, Chiyuan},
  journal={arXiv preprint arXiv:2407.06460},
  year={2024}
}

@article{shi2023detecting,
  title={Detecting pretraining data from large language models},
  author={Shi, Weijia and Ajith, Anirudh and Xia, Mengzhou and Huang, Yangsibo and Liu, Daogao and Blevins, Terra and Chen, Danqi and Zettlemoyer, Luke},
  journal={arXiv preprint arXiv:2310.16789},
  year={2023}
}

@inproceedings{murakonda2021quantifying,
  title={Quantifying the privacy risks of learning high-dimensional graphical models},
  author={Murakonda, Sasi Kumar and Shokri, Reza and Theodorakopoulos, George},
  booktitle={International Conference on Artificial Intelligence and Statistics},
  pages={2287--2295},
  year={2021},
  organization={PMLR}
}

@inproceedings{ye2022enhanced,
  title={Enhanced membership inference attacks against machine learning models},
  author={Ye, Jiayuan and Maddi, Aadyaa and Murakonda, Sasi Kumar and Bindschaedler, Vincent and Shokri, Reza},
  booktitle={Proceedings of the 2022 ACM SIGSAC conference on computer and communications security},
  pages={3093--3106},
  year={2022}
}

@article{ji2024reversing,
  title={Reversing the forget-retain objectives: An efficient llm unlearning framework from logit difference},
  author={Ji, Jiabao and Liu, Yujian and Zhang, Yang and Liu, Gaowen and Kompella, Ramana R and Liu, Sijia and Chang, Shiyu},
  journal={Advances in Neural Information Processing Systems},
  volume={37},
  pages={12581--12611},
  year={2024}
}

@article{karamolegkou2023copyright,
  title={Copyright violations and large language models},
  author={Karamolegkou, Antonia and Li, Jiaang and Zhou, Li and S{\o}gaard, Anders},
  journal={arXiv preprint arXiv:2310.13771},
  year={2023}
}

@article{patil2023can,
  title={Can sensitive information be deleted from llms? objectives for defending against extraction attacks},
  author={Patil, Vaidehi and Hase, Peter and Bansal, Mohit},
  journal={arXiv preprint arXiv:2309.17410},
  year={2023}
}

@inproceedings{ye2025towards,
  title={Towards safe machine unlearning: A paradigm that mitigates performance degradation},
  author={Ye, Shanshan and Lu, Jie and Zhang, Guangquan},
  booktitle={Proceedings of the ACM on Web Conference 2025},
  pages={4635--4652},
  year={2025}
}

@article{achiam2023gpt,
  title={Gpt-4 technical report},
  author={Achiam, Josh and Adler, Steven and Agarwal, Sandhini and Ahmad, Lama and Akkaya, Ilge and Aleman, Florencia Leoni and Almeida, Diogo and Altenschmidt, Janko and Altman, Sam and Anadkat, Shyamal and others},
  journal={arXiv preprint arXiv:2303.08774},
  year={2023}
}

@article{bai2023qwen,
  title={Qwen technical report},
  author={Bai, Jinze and Bai, Shuai and Chu, Yunfei and Cui, Zeyu and Dang, Kai and Deng, Xiaodong and Fan, Yang and Ge, Wenbin and Han, Yu and Huang, Fei and others},
  journal={arXiv preprint arXiv:2309.16609},
  year={2023}
}

@article{jia2024soul,
  title={Soul: Unlocking the power of second-order optimization for llm unlearning},
  author={Jia, Jinghan and Zhang, Yihua and Zhang, Yimeng and Liu, Jiancheng and Runwal, Bharat and Diffenderfer, James and Kailkhura, Bhavya and Liu, Sijia},
  journal={arXiv preprint arXiv:2404.18239},
  year={2024}
}

@article{yao2024machine,
  title={Machine unlearning of pre-trained large language models},
  author={Yao, Jin and Chien, Eli and Du, Minxin and Niu, Xinyao and Wang, Tianhao and Cheng, Zezhou and Yue, Xiang},
  journal={arXiv preprint arXiv:2402.15159},
  year={2024}
}

@article{liu2025rethinking,
  title={Rethinking machine unlearning for large language models},
  author={Liu, Sijia and Yao, Yuanshun and Jia, Jinghan and Casper, Stephen and Baracaldo, Nathalie and Hase, Peter and Yao, Yuguang and Liu, Chris Yuhao and Xu, Xiaojun and Li, Hang and others},
  journal={Nature Machine Intelligence},
  pages={1--14},
  year={2025},
  publisher={Nature Publishing Group UK London}
}

@article{li2024wmdp,
  title={The wmdp benchmark: Measuring and reducing malicious use with unlearning},
  author={Li, Nathaniel and Pan, Alexander and Gopal, Anjali and Yue, Summer and Berrios, Daniel and Gatti, Alice and Li, Justin D and Dombrowski, Ann-Kathrin and Goel, Shashwat and Phan, Long and others},
  journal={arXiv preprint arXiv:2403.03218},
  year={2024}
}

@article{torkzadehmahani2024improved,
  title={Improved localized machine unlearning through the lens of memorization},
  author={Torkzadehmahani, Reihaneh and Nasirigerdeh, Reza and Kaissis, Georgios and Rueckert, Daniel and Dziugaite, Gintare Karolina and Triantafillou, Eleni},
  journal={arXiv preprint arXiv:2412.02432},
  year={2024}
}

@article{brown2021language,
  title={Language models are few-shot learners},
  author={Brown, Tom B and Mann, Benjamin and Ryder, Nick and Subbiah, Melanie and Kaplan, Jared D and Dhariwal, Prafulla and Neelakantan, Arvind and Shyam, Pranav and Sastry, Girish and Askell, Amanda and others},
  journal={Advances in Neural Information Processing Systems},
  volume={33},
  pages={1877--1901},
  year={2021}
}

@inproceedings{hayes2025inexact,
  title={Inexact unlearning needs more careful evaluations to avoid a false sense of privacy},
  author={Hayes, Jamie and Shumailov, Ilia and Triantafillou, Eleni and Khalifa, Amr and Papernot, Nicolas},
  booktitle={2025 IEEE Conference on Secure and Trustworthy Machine Learning (SaTML)},
  pages={497--519},
  year={2025},
  organization={IEEE}
}

@article{pawelczyk2024machine,
  title={Machine unlearning fails to remove data poisoning attacks},
  author={Pawelczyk, Martin and Di, Jimmy Z and Lu, Yiwei and Sekhari, Ayush and Kamath, Gautam and Neel, Seth},
  journal={arXiv preprint arXiv:2406.17216},
  year={2024}
}

@inproceedings{shokri2017membership,
  title={Membership inference attacks against machine learning models},
  author={Shokri, Reza and Stronati, Marco and Song, Congzheng and Shmatikov, Vitaly},
  booktitle={2017 IEEE symposium on security and privacy (SP)},
  pages={3--18},
  year={2017},
  organization={IEEE}
}

@inproceedings{chen-etal-2025-safeeraser,
    title = "{S}afe{E}raser: Enhancing Safety in Multimodal Large Language Models through Multimodal Machine Unlearning",
    author = "Chen, Junkai  and
      Deng, Zhijie  and
      Zheng, Kening  and
      Yan, Yibo  and
      Liu, Shuliang  and
      Wu, PeiJun  and
      Jiang, Peijie  and
      Liu, Jia  and
      Hu, Xuming",
    booktitle = "Findings of the Association for Computational Linguistics: ACL 2025",
    month = jul,
    year = "2025",
    pages = "14194--14224",
}

@mastersthesis{zhao2025rethinking,
  title={Rethinking adversarial robustness in the context of the right to be forgotten},
  author={Zhao, Chenxu},
  year={2025},
  school={Iowa State University}
}

@article{lucki2024adversarial,
  title={An adversarial perspective on machine unlearning for ai safety},
  author={{\L}ucki, Jakub and Wei, Boyi and Huang, Yangsibo and Henderson, Peter and Tram{\`e}r, Florian and Rando, Javier},
  journal={arXiv preprint arXiv:2409.18025},
  year={2024}
}

@article{xue2025dual,
  title={Dual-View Inference Attack: Machine Unlearning Amplifies Privacy Exposure},
  author={Xue, Lulu and Hu, Shengshan and Qian, Linqiang and Guo, Peijin and Zhang, Yechao and Li, Minghui and Zhang, Yanjun and Ye, Dayong and Zhang, Leo Yu},
  journal={arXiv preprint arXiv:2512.16126},
  year={2025}
}

@inproceedings{meng2025rr,
  title={Rr: Unveiling llm training privacy through recollection and ranking},
  author={Meng, Wenlong and Zhenyuan, Guo and Wu, Lenan and Gong, Chen and Liu, Wenyan and Li, Weixian and Wei, Chengkun and Chen, Wenzhi},
  booktitle={Findings of the Association for Computational Linguistics: ACL 2025},
  pages={17383--17397},
  year={2025}
}

@article{bao2025module,
  title={Module-Aware Parameter-Efficient Machine Unlearning on Transformers},
  author={Bao, Wenjie and Lou, Jian and Hu, Yuke and Li, Xiaochen and Liu, Zhihao and Liu, Jiaqi and Qin, Zhan and Ren, Kui},
  journal={arXiv preprint arXiv:2508.17233},
  year={2025}
}

@inproceedings{hu2024eraser,
  title={Eraser: Machine unlearning in mlaas via an inference serving-aware approach},
  author={Hu, Yuke and Lou, Jian and Liu, Jiaqi and Ni, Wangze and Lin, Feng and Qin, Zhan and Ren, Kui},
  booktitle={Proceedings of the 2024 on ACM SIGSAC Conference on Computer and Communications Security},
  pages={3883--3897},
  year={2024}
}

@article{liu2026aligned,
  title   = {Aligned but Fragile: Enhancing {LLM} Safety Robustness via Zeroth-Order Optimization},
  author  = {Liu, Zhihao and Wu, Yifan and Lou, Jian and Wang, Di and Zhou, Yuxi and Hu, Yuke},
  journal = {arXiv preprint arXiv:2605.29396},
  year    = {2026}
}

@inproceedings{hu2025vlmmia,
  title     = {Membership Inference Attacks Against Vision-Language Models},
  author    = {Hu, Yuke and Li, Zheng and Liu, Zhihao and Zhang, Yang and Qin, Zhan and Ren, Kui and Chen, Chun},
  booktitle = {34th USENIX Security Symposium (USENIX Security 25)},
  pages     = {1589--1608},
  year      = {2025}
}

@article{miao2025towards,
  title   = {Towards Evaluation for Real-World {LLM} Unlearning},
  author  = {Miao, Ke and Hu, Yuke and Li, Xiaochen and Bao, Wenjie and Liu, Zhihao and Qin, Zhan and Ren, Kui},
  journal = {arXiv preprint arXiv:2508.01324},
  year    = {2025}
}

@article{wang2026deconstruct,
  title   = {Deconstruct Evaluate and Targeted Intervention Based {LLMs} Unlearning Algorithm},
  author  = {Wang, Yang and Miao, Ke and Hu, Yuke and Li, Xiaochen and Qin, Zhan},
  journal = {Chinese Journal of Electronics},
  volume  = {35},
  number  = {2},
  pages   = {761--770},
  year    = {2026},
  doi     = {10.23919/cje.2025.00.105}
}

@article{hu2025fkge,
  title   = {Privacy Risks of Federated Knowledge Graph Embedding: New Membership Inference Attacks and Personalized Differential Privacy Defense},
  author  = {Hu, Yuke and Wang, Yang and Lou, Jian and Liang, Wei and Wu, Ruofan and Wang, Weiqiang and Li, Xiaochen and Liu, Jinfei and Qin, Zhan},
  journal = {IEEE Transactions on Dependable and Secure Computing},
  volume  = {22},
  number  = {3},
  pages   = {2788--2805},
  year    = {2025},
  doi     = {10.1109/TDSC.2024.3522025}
}

@inproceedings{hu2023fkge,
  title     = {Quantifying and Defending against Privacy Threats on Federated Knowledge Graph Embedding},
  author    = {Hu, Yuke and Liang, Wei and Wu, Ruofan and Xiao, Kai and Wang, Weiqiang and Li, Xiaochen and Liu, Jinfei and Qin, Zhan},
  booktitle = {Proceedings of the ACM Web Conference 2023},
  pages     = {2306--2317},
  year      = {2023}
}

@inproceedings{miao2026safe,
  title     = {Adaptive and Explicit {\texttt{<safe>}}: Triggering Latent Safety Awareness in Large Reasoning Models},
  author    = {Miao, Ke and Li, Jiaxin and Chen, Hongliang and Hu, Yuke and Qin, Zhan},
  booktitle = {Joint European Conference on Machine Learning and Knowledge Discovery in Databases (ECML PKDD)},
  pages     = {189--205},
  year      = {2026},
  publisher = {Springer},
  doi       = {10.1007/978-3-032-37673-2_11}
}

@article{liu2026dpzo,
  title   = {Differentially Private Zeroth-Order Methods for Scalable Large Language Model Fine-Tuning},
  author  = {Liu, Zhihao and Lou, Jian and Bao, Wenjie and Hu, Yuke and Wang, Wenli and Qin, Zhan and Ren, Kui},
  journal = {IEEE Transactions on Information Forensics and Security},
  volume  = {21},
  pages   = {4163--4175},
  year    = {2026},
  doi     = {10.1109/TIFS.2026.3679194}
}

\begin{IEEEbiography}
[{\includegraphics[width=1in,height=1.25in,clip,keepaspectratio]{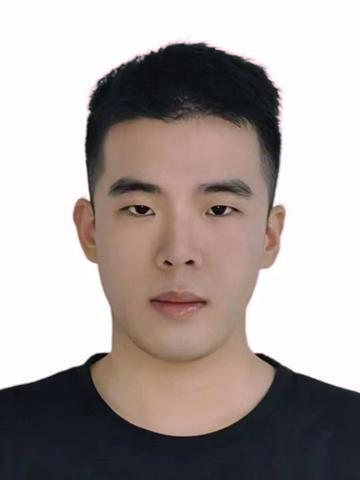}}] Zhihao~Liu received the BS degree in information security from Shandong University. He is working toward the PhD degree with Zhejiang University. His research interests focus on protecting model security and data privacy in large language models, with an emphasis on differential privacy, machine unlearning, and safety alignment.
\end{IEEEbiography}

\begin{IEEEbiography}
[{\includegraphics[width=1in,height=1.25in,clip,keepaspectratio]{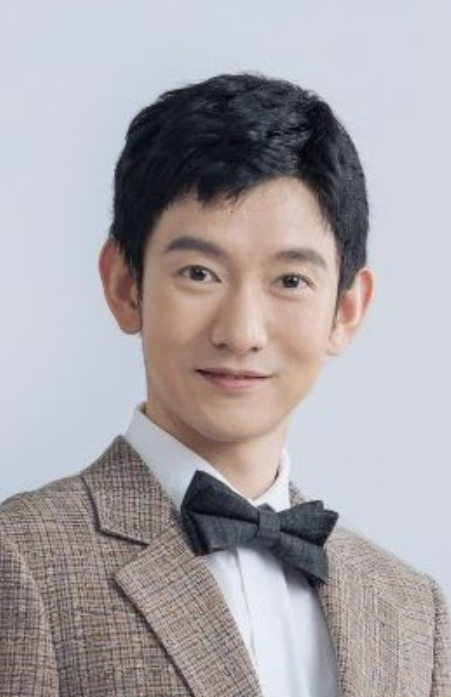}}] Jian Lou is an assistant professor at Sun Yat-sen University. He was a Researcher in ZJU-Hangzhou Global Scientific and Technological Innovation Center from 2022 to 2024, an associate professor at Xidian University from 2021 to 2022. He was a Post-doc in the Department of Computer Science at Emory University from 2019 to 2021 and he obtained his Ph.D. in Computer Science at Hong Kong Baptist University in 2018. His current research interests lie in the Differential Privacy, Machine Unlearning and Certified Robustness.
\end{IEEEbiography}

\begin{IEEEbiography}
[{\includegraphics[width=1in,height=1.25in,clip,keepaspectratio]{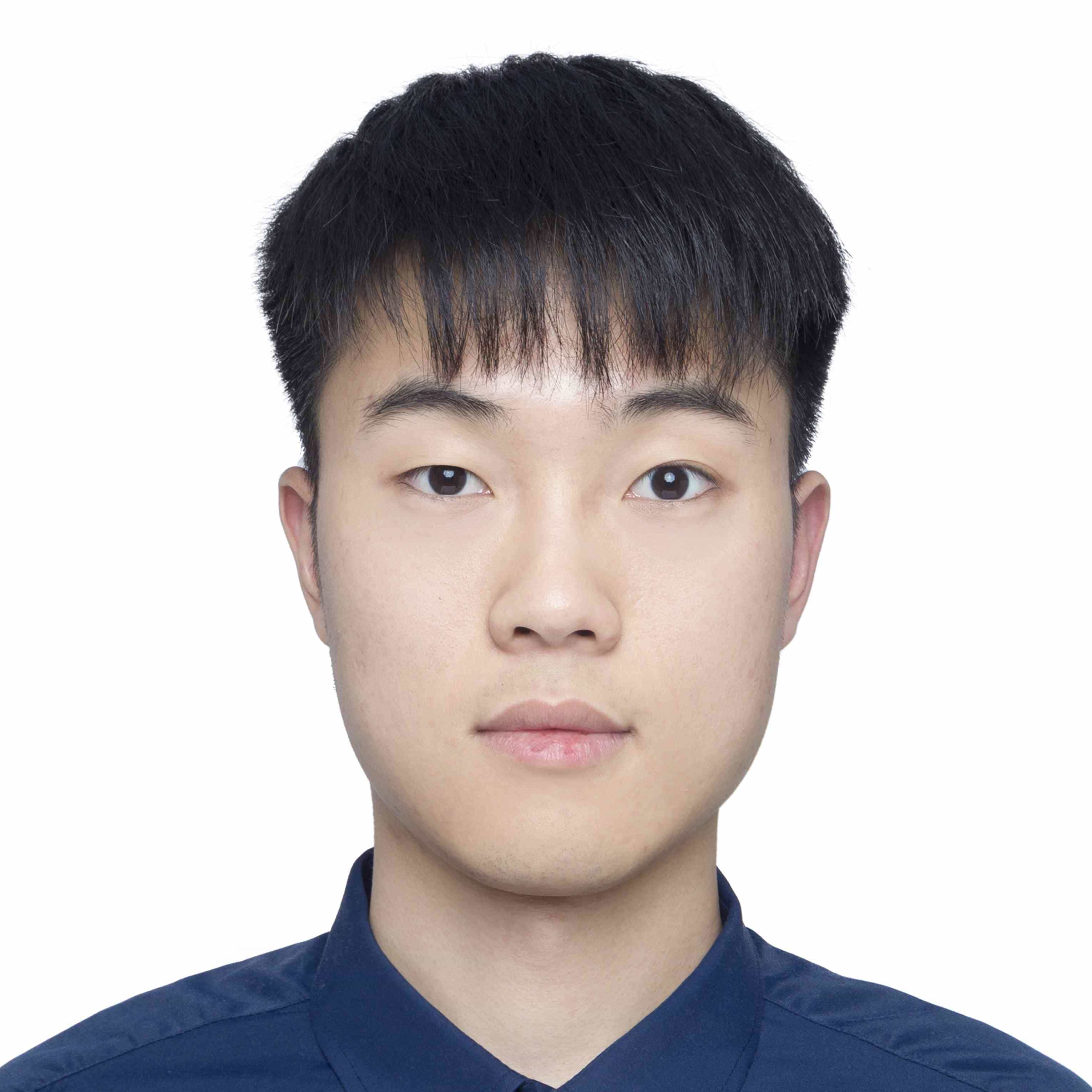}}] Yuke Hu received the bachelor’s degree in information security from Wuhan
University. He is currently working toward the PhD degree with Zhejiang University. His research interests include data security and privacy in multiscenario applications and trustworthy AI.
\end{IEEEbiography}

\begin{IEEEbiography}
[{\includegraphics[width=1in,height=1.25in,clip,keepaspectratio]{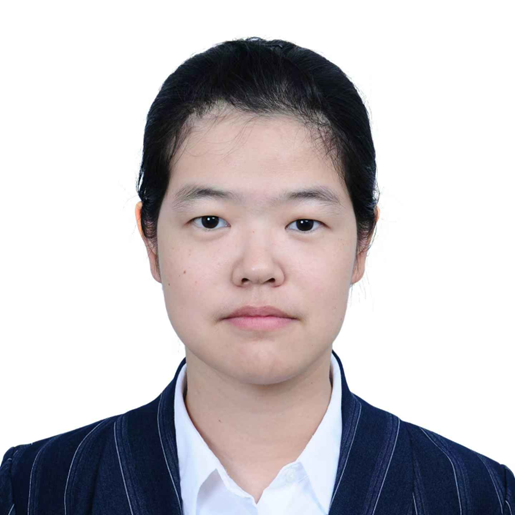}}] Xiaochen Li received her Ph.D. in Computer Science and Technology from Zhejiang University, China, in 2024. She is currently an Assistant Professor in the Department of Computer Science at the University of North Carolina at Greensboro, USA. Prior to this, she was a Postdoctoral Researcher in the Department of Computer Science at the University of Virginia, USA. Her research focuses on data privacy, with a particular emphasis on differential privacy. Her work has been published in leading conferences and journals, including IEEE S\&P, NDSS, VLDB, SIGMOD, and IEEE TDSC.
\end{IEEEbiography}

\begin{IEEEbiography}
[{\includegraphics[width=1in,height=1.25in,clip,keepaspectratio]{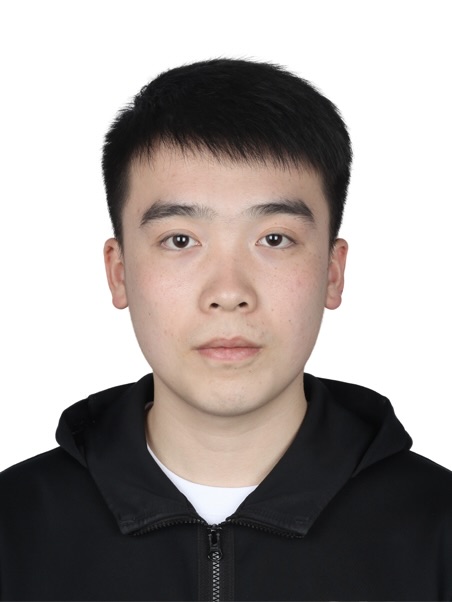}}] Yitian Cheng is a master’s student at Zhejiang University. After completing his B.S. in Computer Science at Xi’an Jiaotong University, he has been focusing on research topics such as differential privacy and machine unlearning.
\end{IEEEbiography}

\begin{IEEEbiography}
[{\includegraphics[width=1in,height=1.25in,clip,keepaspectratio]{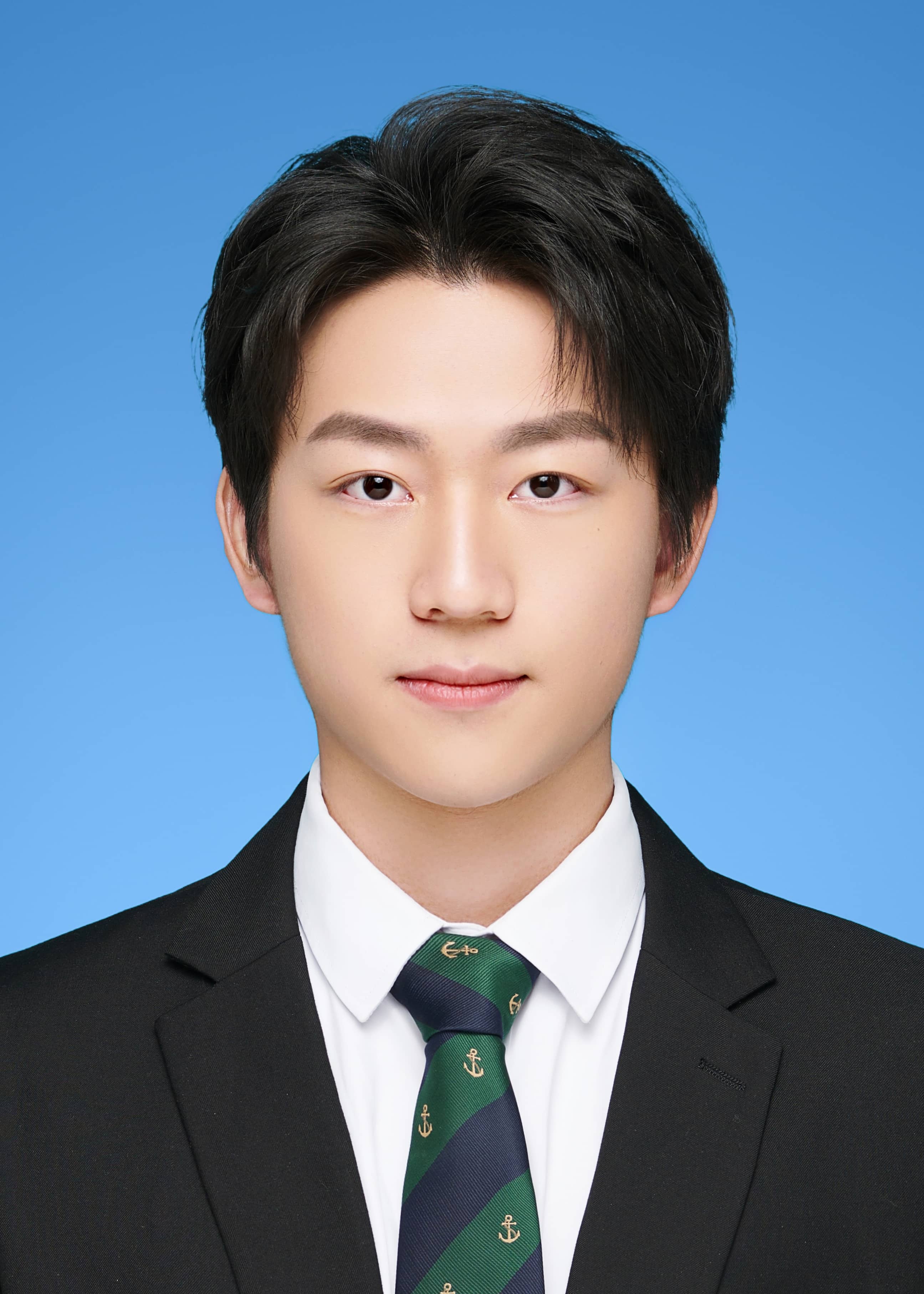}}] Tailun Chen received the B.S. degree in information security from the Nanjing University of Aeronautics and Astronautics. He is currently pursuing the Ph.D. degree with Zhejiang University. His research interests include security of large language models and AI agents.
\end{IEEEbiography}

\begin{IEEEbiography}
[{\includegraphics[width=1in,height=1.25in,clip,keepaspectratio]{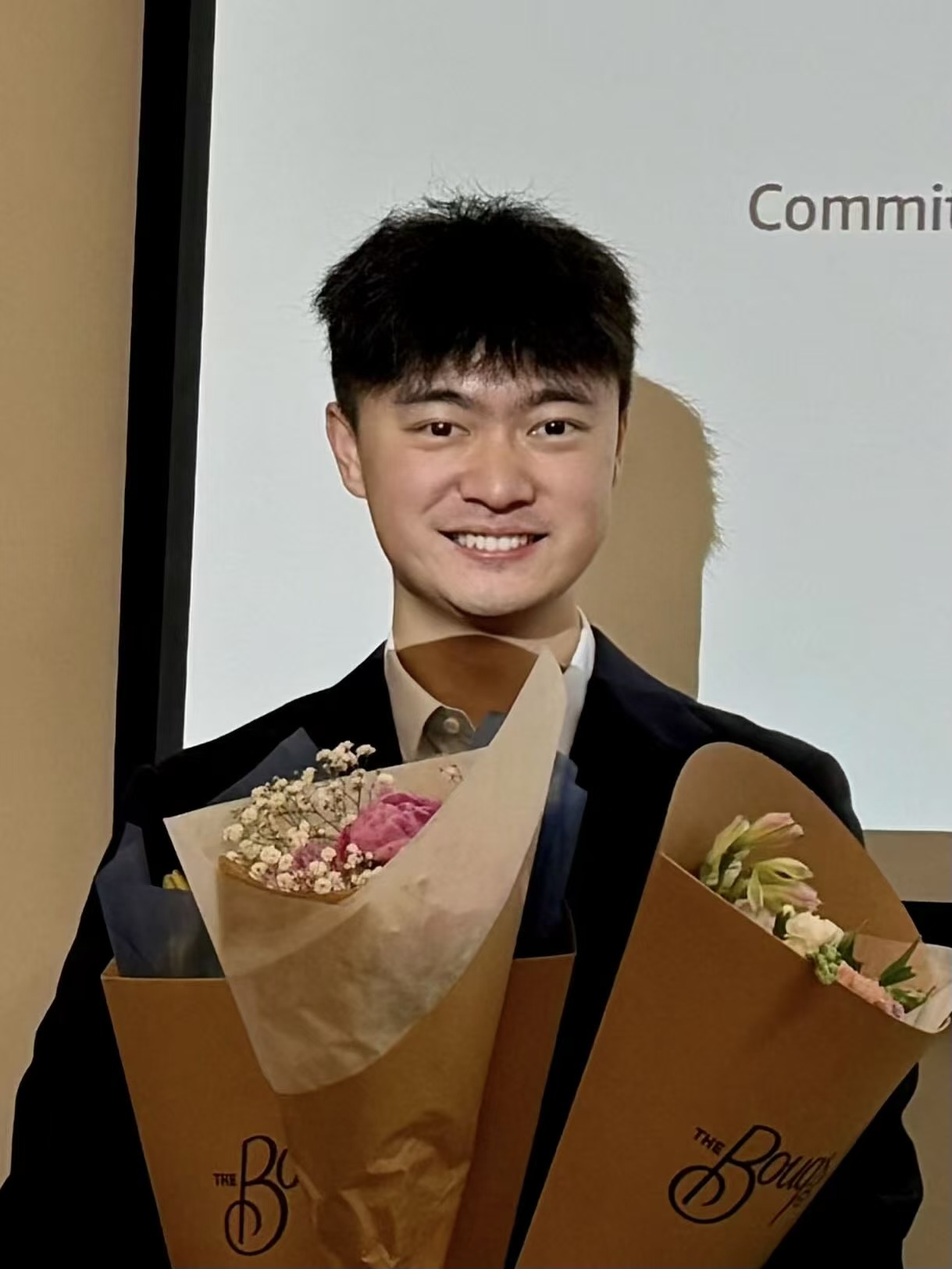}}] Zhizhen Qin received the Ph.D. degree in Computer Science from the University of California, San Diego. He is currently an Applied Scientist at Amazon. He has authored papers accepted to top-tier AI conferences including NeurIPS, AAAI, and ICRA. He has also co-organized Math AI workshops at ICML 2025 and NeurIPS 2025. His research interests include learning-based verification, formal reasoning with large language models, and safety and robustness of learning-based control systems.
\end{IEEEbiography}

\begin{IEEEbiography}
[{\includegraphics[width=1in,height=1.25in,clip,keepaspectratio]{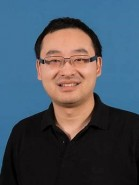}}] Zhan Qin received the PhD degree in the computer science and Engineering Department from the State University of New York, Buffalo, in 2017. He is currently an associate professor, with both the College of Computer Science and Technology and the Institute of Cyberspace Research (ICSR), Zhejiang University, China. He was an assistant professor with the Department of Electrical and Computer Engineering, University of Texas, San Antonio. His current research interests include data security and privacy, secure computation outsourcing, artificial intelligence security, and cyberphysical security in the context of the Internet of Things. His works explore and develop novel security sensitive algorithms and protocols for computation and communication on the general context of Cloud and Internet devices.
\end{IEEEbiography}

\begin{IEEEbiography}
[{\includegraphics[width=1in,height=1.25in,clip,keepaspectratio]{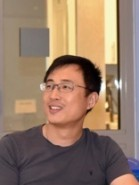}}] Kui Ren received degrees from three different majors, i.e., his PhD in electrical and computer engineering from Worcester Polytechnic Institute, USA, in 2007, the MEng in materials engineering, in 2001, and BEng degree in chemical engineering, in 1998, both from Zhejiang University, China. He is currently a professor and associate Dean of College of Computer Science and Technology with Zhejiang University, where he also directs the Institute of Cyber Science and Technology. Kui’s current research interests include Data Security, IoT Security, AI Security, and Privacy.
\end{IEEEbiography}

\clearpage
\appendix
\section*{Open Science}
Our research team is committed to the principles of open science, making our findings freely accessible. The model and dataset we use in our experiments can be downloaded on HuggingFace. We hope that releasing our code as open source will not only support further research but also potentially influence real-world applications in the industry. The code is available at \href{https://github.com/Liuxiaohao6/EGU-unlearning}{https://github.com/Liuxiaohao6/EGU-unlearning}.

\section{Appendix}

\subsection{Hardware configuration}
We conduct all experiments on 4 A6000-48G GPUs. All experiments are conducted with and torch 2.2 and CUDA 12.1. We employ flash-attention-2 2.5.7 to improve the training and inference efficiency. We employ DeepSpeed ZeRO stage-3 for all baselines to compress GPU memory following the previous implementation released by TOFU \cite{maini2024tofu}.

\subsection{More experiment details and Details of metrics}\label{sec:muse metric}
In this section, we list more details of unlearning experiment and the details of how to calculate the metrics we use.

For each proxy generation prompt, we use a small set of few-shot demonstrations, each containing the original question, the original answer, and a corresponding \emph{pure answer}. In our experiments, the \emph{pure answer} is produced by the retrained model, serving as a clean reference for evaluating whether proxy answers approximate the outcome of retraining-based unlearning. This use of the retrained model is only for controlled analysis and is not required in deployment. In practice, the demonstrations can be collected from previously successful unlearning cases, whose post-unlearning answers can be reused to guide future proxy generation. Thus, EGUP does not require repeated retraining for new unlearning requests.

Each model on MUSE is trained for 10 epochs with a constant learning rate 1e-5 and a batch size of 16. Results are selected from 10 checkpoints saved at each epoch. The unlearning epochs for each unlearning method on MUSE is presented in Table \ref{tab:muse epoch} with the same learning rate of 1e-5.
\begin{table}[ht]
\centering
\setlength{\tabcolsep}{15pt} 
\caption{Unlearning epochs for each unlearning method on MUSE.}\label{tab:muse epoch}
\begin{tabular}{@{}lcc@{}}
\toprule
\textbf{Unlearning Method} & \textbf{News} & \textbf{Books} \\
\midrule
GA               & epoch 1     & epoch 1     \\
EGUP(GA)     & epoch 1     & epoch 1     \\
GD               & epoch 7     & epoch 1     \\
EGUP(GD)     & epoch 7     & epoch 1     \\
NPO              & epoch 1     & epoch 1     \\
EGUP(NPO)    & epoch 1     & epoch 1     \\
NPO+GD           & epoch 10    & epoch 1     \\
EGUP(NPO+GD) & epoch 10    & epoch 1     \\
\bottomrule
\end{tabular}
\end{table}

\subsubsection{Experiment Setups of TOFU}
We fine-tune both the Phi and LLaMA models using identical hyper-parameters. Specifically, we adopt a batch size of 32, gradient accumulation steps of 1, a total of 10 training epochs, a learning rate of $1 \times 10^{-5}$, weight decay of 0.01, and a fixed random seed of 42. In the subsequent unlearning phase, model-specific configurations are applied. For the Phi model, we use a batch size of 16 and train for only 1 epoch with gradient accumulation steps remaining at 1. The retain strength coefficient $\alpha$ is fixed at 1 throughout. When applying the NPO method, the inverse temperature is set to $\beta = 2.5$. In contrast, for our proposed entanglement-aware framework, we perform a grid search over entanglement temperatures $\{0.5,\ 1,\ 2\}$, and set the penalty weight $\mu$ through a grid search over $\{0.001,\ 0.005\}$. 
For the LLaMA model, we adopt a smaller batch size of 8 to accommodate its larger memory footprint. The number of training epochs is set to 2, while the gradient accumulation steps remain consistent with those used for the Phi model. Additionally, we apply LoRA during unlearning, with rank $r=32$, $\alpha=32$, and dropout of 0.05. Weight decay is also set to 0.01. All other unlearning-related hyper-parameters, including $\alpha$, $\beta$, the entanglement temperature, and $\mu$ are the same as in the Phi configuration.

\subsubsection{Details of Unlearning Objectives}
\label{app:unlearning_objectives}

This appendix provides the detailed formulations of the unlearning objectives used in our experiments.

Unlearning objectives for fine-tuning target LLMs include gradient-ascent methods and preference-loss methods. Notable examples of the forget loss and retain loss are as follows:
\begin{itemize}[leftmargin=*, labelindent=0.2em]
    \item Gradient Ascent (GA): The forget loss is formulated as:
    \begin{equation}
        L_{\text{GA}}(\theta',\mathcal{D}_f)=-\mathbb{E}_{\mathcal{D}_f}[\log(f(\theta',y|x))],
    \end{equation}
    where the optimization is performed on the forget dataset $\mathcal{D}_f$ without referencing the retain dataset $\mathcal{D}_r$. Notably, the target label $y$ used in this objective can be flexibly chosen. Instead of the original correct output, one may substitute it with an arbitrary response $\widetilde{y}$ that the model is encouraged to output post-unlearning, such as random answers \cite{eldan2023s}, or simply answering "I don't know" \cite{maini2024tofu}.
    \item Gradient Difference (GD): The forget loss is the same as GA, and the retain loss is formulated as:
    \begin{equation}
        L(\theta',\mathcal{D}_r)=-\mathbb{E}_{\mathcal{D}_r}[\log(f(\theta',y|x))],
    \end{equation}
    where $(x,y)\in\mathcal{D}_r$, which encourages the model to still perform well on the retain dataset $\mathcal{D}_r$.
    \item Negative Preference Optimization (NPO) \cite{zhang2024negative}: NPO performs preference optimization using only negative responses as supervision signals. The forget loss is formulated as:
    \begin{small}
        \begin{equation}
        L_{\text{NPO},\beta}(\theta')=-\frac{2}{\beta}\mathbb{E}_{\mathcal{D}_f}\left[\log\sigma
        (-\beta\log\frac{f(\theta',y|x)}{f(\theta^{o},y|x)})\right],
        \end{equation}
    \end{small}
    
    where $\sigma(t)=1/(1+e^{-t})$ is the sigmoid function, $\beta>0$ is the inverse temperature and $f(\theta^{o})$ is the model trained on $\mathcal{D}_{\text{train}}$. NPO+GD combines the NPO loss with the retain loss from GD.
\end{itemize}

\subsubsection{Metric of TOFU Dataset}

We mainly adopt the metric proposed in the original TOFU paper~\cite{maini2024tofu}. 

\textbf{Model Utility.} Model utility is the aggregated metrics across multiple retain sets, including the data of remaining fictional writers other than the authors in forget data, the QA pairs of real-world writers, and general world facts. The model utility is defined as the harmonic average of three metrics evaluated on the aforementioned three groups of retain data, i.e., aggregated value of nine metrics. The metrics include ROUGE-L score between unlearned LLM generated response and ground-truth response, the accuracy of unlearned LLM accuracy on the data, and the average truth-ratio, which is defined by:
\[
R_{\text{truth}} := \frac{1}{N} \sum_{i=1}^N \frac{p(\hat{y}_i|x)^{(1/|\hat{y}_i|)}}{p(\tilde{y}_i|x)^{(1/|\tilde{y}_i|)}},
\]
where $x, \tilde{y}, \hat{y}$ are original questions, incorrect answers, and paraphrased correct answers, respectively, and $N$ is the number of incorrect answers. The rationale of the truth ratio is that it measures how likely the unlearned LLM will give a correct answer versus an incorrect one.

\textbf{Forget Quality.} Forget quality assesses how well the unlearned LLM mimics a retrain LLM, which is trained without the forget data. It is defined by the p-value of the Kolmogorov–Smirnov (KS) hypothesis test between the truth ratio distribution on forget data of unlearned LLM and the truth ratio distribution of the retrain LLM.
We refer readers to the original TOFU paper~\cite{maini2024tofu} for more details.

\subsubsection{Metric of MUSE Dataset} We mainly adopt the metric proposed in the original MUSE paper~\cite{shi2024muse}

\textbf{Verbatim Memorization (VerbMem).}
Considering a model $\theta$ with the first $l$ tokens from a sample $l$, the ROUGE-L value between the output of $f(x_{[l]}; \theta)$ and the true continuation $x_{[l+1:]}$ is defined as the VerbMem:
\begin{equation}
\text{VerbMem}(\theta, \mathcal{D}_u) = \frac{1}{|\mathcal{D}_u|} \sum_{x \in \mathcal{D}_u} \text{ROUGE}(f(x_{[l]}; \theta), x_{[l+1:]}).
\end{equation}

\textbf{Knowledge Memorization (KnowMem).}
Different from VerbMem, KnowMem considers from an instance-wise perspective. Let $(q, a)$ represent a pair of question and answer in one sample $x$, the KnowMem is defined as:
\begin{equation}
\text{KnowMem}(\theta, \mathcal{D}_u) = \frac{1}{|\mathcal{D}_u|} \sum_{(q,a) \in \mathcal{D}_u} \text{ROUGE}(f(q; \theta), a).
\end{equation}

\textbf{Privacy Leakage (PrivLeak).}
It is essential that the unlearned model does not leak membership details that could reveal $\mathcal{D}_u$ is part of $\mathcal{D}_{tr}$. Therefore, to accurately measure the degree of leakage, PrivLeak is proposed by employing Min-K\% Prob \cite{shi2023detecting} and computing the standard AUC-ROC score \cite{murakonda2021quantifying, ye2022enhanced} as follows:
\begin{equation}
\text{PrivLeak} := \frac{\text{AUC}(\theta_r; \mathcal{D}_u, \mathcal{D}_h) - \text{AUC}(\theta_t; \mathcal{D}_u, \mathcal{D}_h)}{\text{AUC}(\theta_r; \mathcal{D}_u, \mathcal{D}_h)},
\end{equation}
where $\theta_r$ is the retrained model on the retain dataset, $\theta_t$ is the unlearned model. $\mathcal{D}_h$ denotes a ‘holdout’ dataset which contains in-distribution samples but the model has not been trained on. A well-performing unlearning algorithm should have a PrivLeak metric close to zero, whereas an over- or under-unlearning algorithm will result in a high positive or negative value.

\textbf{Utility Preservation (UtilPres).}
Except for metrics on unlearn tasks, performance on other tasks are also concerned in \textbf{MUSE}. Specifically, UtilPres is represented as the evaluation on retain dataset with the KnowMem metric.
\begin{equation}
\text{UtilPres}(\theta, \mathcal{D}_r) = \frac{1}{|\mathcal{D}_r|} \sum_{(q,a) \in \mathcal{D}_r} \text{ROUGE}(f(q; \theta), a).
\end{equation}

\subsubsection{Additional Experiments}
Additional experimental results are presented in Table~\ref{tab:phi_tofu1} \ref{tab:phi_tofu5} \ref{tab:phi_tofu10} and Table~\ref{tab:llama_tofu1} \ref{tab:llama_tofu5} \ref{tab:llama_10}, corresponding to evaluations on the TOFU dataset using the Phi and LLaMA models, respectively. Results under varying entanglement temperature $k$ are also reported in the Table~\ref{tab:varyk_phi} and Table~\ref{tab:varyk_llama}.

\begin{figure}[htbp]
    \centering
    \includegraphics[width=\linewidth]{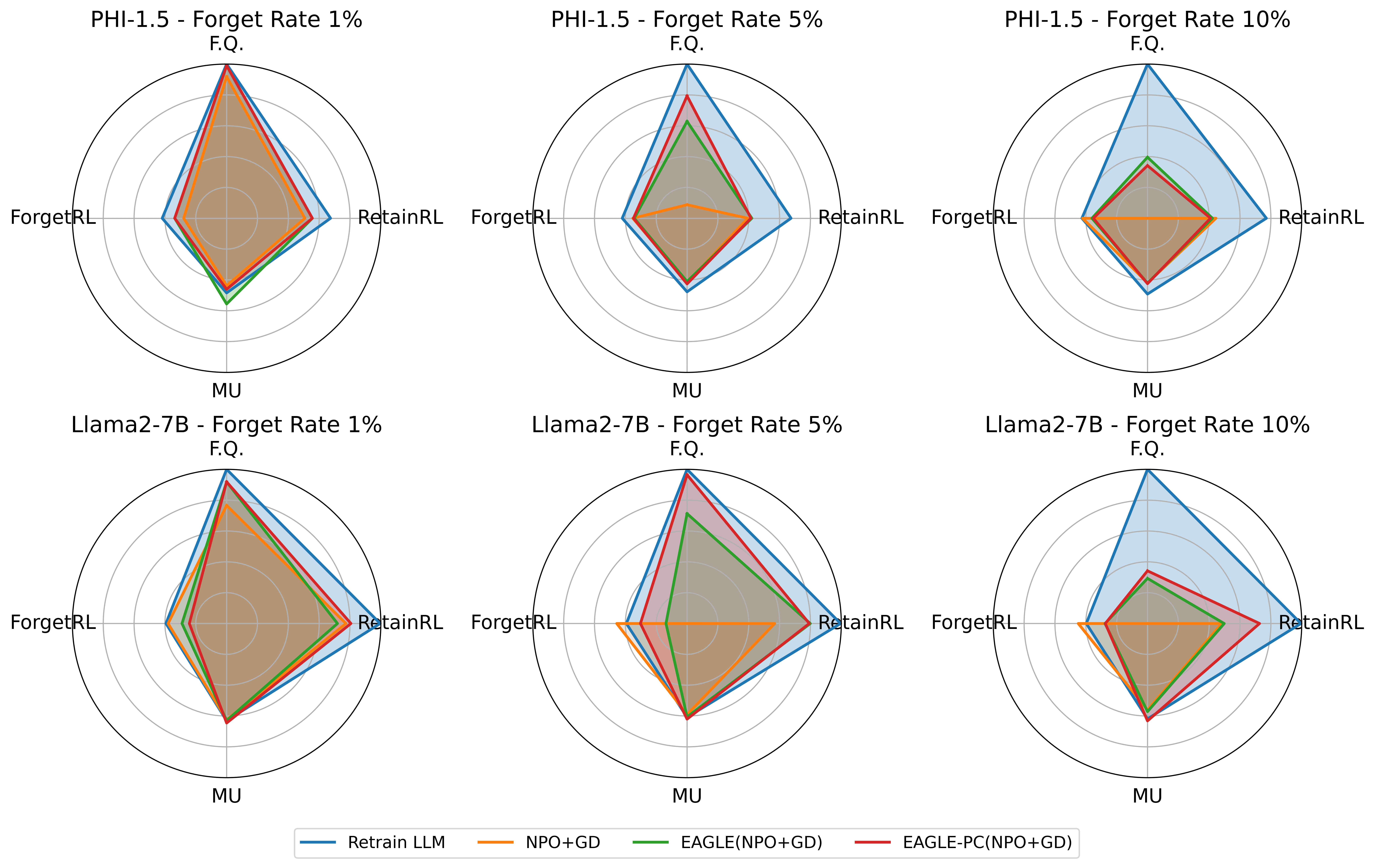}
    \caption{Performance comparison across four key metrics under three forgetting rates. EGUP(NPO+GD) consistently outperforms the other methods across all metrics.}
    \label{fig:leida}
\end{figure}

Figure \ref{fig:leida} compares EGUP with baseline approaches on PHI-1.5 and LLaMA2-7B under different forgetting rates (1\%, 5\%, and 10\%). The radar plots evaluate four key metrics: Forget Quality, ForgetRL, Model Utility, and RetainRL. The blue curve denotes the retrain baseline, serving as the upper-bound target, while the orange curve represents the standard NPO+GD method. Introducing entanglement-guided loss reweighting (green) consistently enhances Forget Quality across all settings. Further incorporating the proxy constraint (red) yields notable improvements in Model Utility and brings performance closer to the retrain upper bound across most metrics. Particularly under the 1\% forgetting scenario, the red curve almost overlaps with the retrain baseline, indicating that EGUP can closely emulate retraining when the forget ratio is small. As the forget rate increases to 5\% and 10\%, the performance gap gradually widens, reflecting the increasing difficulty of jointly maintaining forget quality and model utility.

\begin{figure}[htbp]
    \centering
    \includegraphics[width=\linewidth]{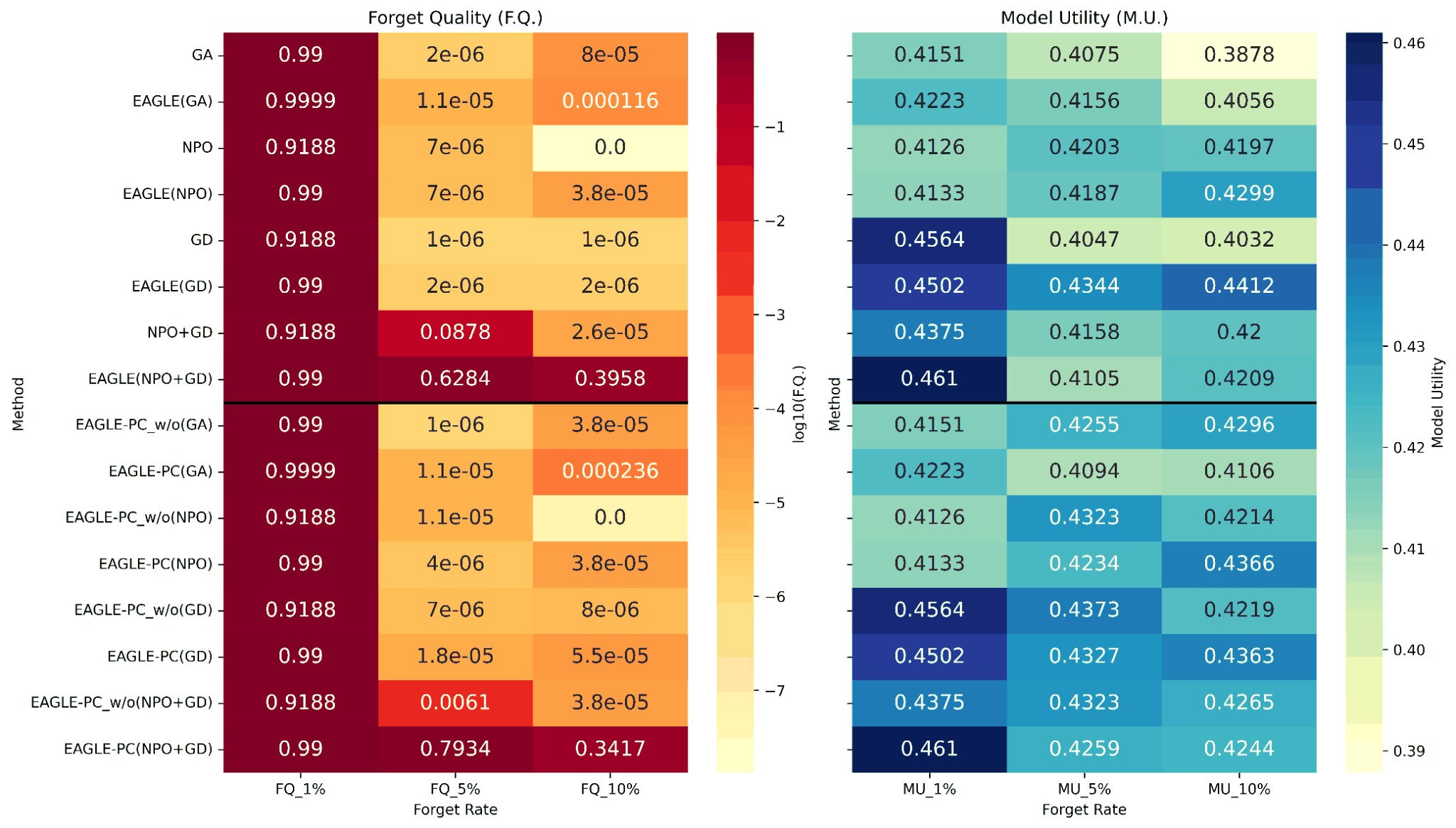}
    \caption{Performance comparison of different unlearning strategies across 1\%, 5\%, 10\% in terms of Forget Quality (left) and Model Utility (right). Black line separates variants without/with proxy constraints. Incorporating test constraint improves utility while maintaining forget quality.}
    \label{fig:hot-test}
\end{figure}

\begin{table}[h]
\centering
\small
\vspace{1em}
\begin{tabular}{lccc}
\toprule
Metric & Mean & Last-128 & Chunk-128 \\
\midrule
VerbMem-F $\downarrow$ & 7.8129 & 7.8129 & 7.8129 \\
PrivLeak $\downarrow$ & 35.5122 & 35.6222 & 35.8622 \\
KnowMem-F $\downarrow$ & 17.7289 & 17.3489 & 16.4389 \\
KnowMem-R $\uparrow$ & 38.8175 & 38.9775 & 39.0175 \\
\bottomrule
\end{tabular}
\vspace{1em}
\caption{Pooling ablation on MUSE BOOKS.}
\label{tab:pooling_ablation}
\end{table}

We compare several unlearning strategies across varying forget rates in Figure \ref{fig:hot-test}. The upper half of each heatmap displays results from baseline and entanglement-guided unlearning (EGU), while the lower half includes EGU with proxy constraint (EGUP). Across all model configurations and forgetting ratios, EGUP consistently achieves better utility preservation while maintaining comparable or even stronger forgetting performance. This demonstrates that incorporating ICL-based proxy guidance provides a more stable and well-regularized optimization trajectory.

We further analyze the limitation of mean pooling on long-context examples. Mean pooling uniformly averages token embeddings to obtain a sentence-level representation, which is simple and stable for short QA pairs but may dilute localized semantic evidence in long passages. This issue is particularly relevant for the BOOKS split in MUSE, where the memorized span is often sparse within a lengthy sequence. To quantify this effect, we compare short QA-style examples with BOOKS verbatim memorization examples. The average sequence length is \textbf{31.5 tokens} for short QA examples, but \textbf{1144.7 tokens} for BOOKS verbatim memorization examples. Under mean pooling, the maximum representation mass contributed by a 10-token critical span drops from \textbf{34.0\%} on short examples to only \textbf{0.90\%} on long examples. Moreover, the critical-span enrichment ratio is close to 1.0 on long examples, meaning that the critical span receives almost the same weight as a random span of the same length. This confirms that mean pooling becomes nearly uniform on long sequences and fails to selectively emphasize semantically critical spans.

We further evaluate whether simple heuristic pooling alternatives can mitigate this limitation. Specifically, we compare mean pooling, last-128-token pooling, and chunk-128 pooling on MUSE BOOKS under our WebA unlearning objective. As shown in Table~\ref{tab:pooling_ablation}, the three pooling strategies achieve nearly identical verbatim forgetting. Chunk-128 pooling slightly improves retained knowledge compared with mean pooling, increasing KnowMem-R from 38.8175 to 39.0175, and reduces KnowMem-F from 17.7289 to 16.4389. However, the overall differences remain small, suggesting that simply replacing mean pooling with a fixed pooling rule is insufficient to fully address the long-context limitation. These results support the reviewer's concern and indicate that future extensions should explore learned span-aware or attention-based pooling mechanisms for long-context unlearning.

\begin{table}[t]
\centering
\small
\caption{Ablation study of EGU and proxy constraint on MUSE NEWS.}
\label{tab:ablation_news}
\renewcommand{\arraystretch}{1.18}
\setlength{\tabcolsep}{3.5pt}
\begin{tabularx}{0.95\linewidth}{lYYYY}
\toprule
\textbf{Method} 
& \textbf{VerbMem} $\downarrow$
& \textbf{KnowMem-F} $\downarrow$
& \textbf{PrivLeak} $\rightarrow 0$
& \textbf{KnowMem-R} $\uparrow$ \\
\midrule
NPO+GD & 16.3903 & 37.1143 & -97.7578 & 37.4993 \\
+ EGU & 3.8427 & 31.2864 & 76.5182 & 30.2146 \\
+ Proxy & 13.7054 & 36.4921 & -86.3417 & 44.8362 \\
\rowcolor{cyan!10}
EGUP & 5.6201 & 35.8123 & 83.2492 & 46.4921 \\
\bottomrule
\end{tabularx}
\vspace{1em}
\end{table}

\begin{table}[t]
\centering
\small
\caption{Ablation study of EGU and proxy constraint on MUSE BOOKS.}
\label{tab:ablation_books}
\renewcommand{\arraystretch}{1.18}
\setlength{\tabcolsep}{3.5pt}
\begin{tabularx}{0.95\linewidth}{lYYYY}
\toprule
\textbf{Method} 
& \textbf{VerbMem} $\downarrow$
& \textbf{KnowMem-F} $\downarrow$
& \textbf{PrivLeak} $\rightarrow 0$
& \textbf{KnowMem-R} $\uparrow$ \\
\midrule
NPO+GD & 26.6586 & 26.4127 & -53.4026 & 43.8690 \\
+ EGU & 4.9368 & 13.5274 & -27.1843 & 31.6428 \\
+ Proxy & 19.3845 & 23.9067 & -47.2158 & 46.3804 \\
\rowcolor{cyan!10}
EGUP & 7.8129 & 17.7289 & -35.5122 & 38.8175 \\
\bottomrule
\end{tabularx}
\end{table}


To further understand the contribution of each component, we conduct an ablation study by separately adding EGU and the proxy constraint to the base NPO+GD objective. As shown in Tables~\ref{tab:ablation_news} and~\ref{tab:ablation_books}, EGU mainly strengthens the unlearning side of the trade-off. For example, on MUSE BOOKS, adding EGU reduces VerbMem from 26.6586 to 4.9368 and KnowMem-F from 26.4127 to 13.5274, indicating stronger removal of verbatim and knowledge memorization. However, this more aggressive unlearning also decreases KnowMem-R from 43.8690 to 31.6428, suggesting a higher risk of over-unlearning.

The proxy constraint plays a complementary role. When used alone, it better preserves retained utility but provides weaker forgetting improvement than EGU. For instance, on MUSE BOOKS, +Proxy improves KnowMem-R from 43.8690 to 46.3804, but only reduces VerbMem from 26.6586 to 19.3845. When combined with EGU, EGUP achieves a more balanced trade-off: it keeps VerbMem and KnowMem-F substantially lower than the vanilla objective, while recovering retained utility compared with EGU-only. On MUSE BOOKS, EGUP improves KnowMem-R from 31.6428 under EGU-only to 38.8175, while still reducing VerbMem from 26.6586 to 7.8129 and KnowMem-F from 26.4127 to 17.7289. These results show that EGU strengthens unlearning, whereas the proxy constraint mitigates the utility degradation caused by overly aggressive unlearning.

\clearpage

\begin{table}[t]
\centering
\caption{PHI Results on TOFU-1\% benchmark.}
\label{tab:phi_tofu1}
\resizebox{0.7\linewidth}{!}{
\begin{tabular}{l|cc|cc}
\toprule
\multirow{2}{*}{\textbf{Method}} & \multicolumn{2}{c|}{\textbf{Forget Perf.}} & \multicolumn{2}{c}{\textbf{Retain Perf.}} \\
\cmidrule{2-5}
& F.Q. $\uparrow$ & R-L $\uparrow$ & M.U. $\uparrow$ & R-L $\uparrow$ \\
\midrule
Retrain LLM & 1 & 0.4176 & 0.4845 & 0.6737 \\
\midrule
GA & 0.9900 & 0.2914 & 0.4151 & 0.4722 \\
EGUP\_w/o(GA) & 0.9900 & 0.2914 & 0.4151 & 0.4722 \\
EGU(GA) & 0.9999 & 0.3260 & 0.4223 & 0.4843 \\
EGUP(GA) & 0.9999 & 0.3260 & 0.4223 & 0.4843 \\
\midrule
NPO & 0.9188 & 0.2696 & 0.4126 & 0.4570 \\
EGUP\_w/o(NPO) & 0.9188 & 0.2696 & 0.4126 & 0.4570 \\
EGU(NPO) & 0.9900 & 0.3016 & 0.4133 & 0.4762 \\
EGUP(NPO) & 0.9900 & 0.3016 & 0.4133 & 0.4762 \\
\midrule
GD & 0.9188 & 0.3193 & 0.4564 & 0.5456 \\
EGUP\_w/o(GD) & 0.9188 & 0.3193 & 0.4564 & 0.5456 \\
EGU(GD) & 0.9900 & 0.3594 & 0.4502 & 0.5564 \\
EGUP(GD) & 0.9900 & 0.3594 & 0.4502 & 0.5564 \\
\midrule
NPO+GD & 0.9188 & 0.2783 & 0.4375 & 0.5072 \\
EGUP\_w/o(NPO+GD) & 0.9188 & 0.2783 & 0.4375 & 0.5072 \\
EGU(NPO+GD) & 0.9900 & 0.3366 & 0.4610 & 0.5558 \\
EGUP(NPO+GD) & 0.9900 & 0.3366 & 0.4610 & 0.5558 \\
\bottomrule
\end{tabular}
}
\end{table}

\begin{table}[t]
\centering
\caption{PHI Results on TOFU-5\% benchmark.}
\label{tab:phi_tofu5}
\resizebox{0.7\linewidth}{!}{
\begin{tabular}{l|cc|cc}
\toprule
\multirow{2}{*}{\textbf{Method}} & \multicolumn{2}{c|}{\textbf{Forget Perf.}} & \multicolumn{2}{c}{\textbf{Retain Perf.}} \\
\cmidrule{2-5}
& F.Q. $\uparrow$ & R-L $\uparrow$ & M.U. $\uparrow$ & R-L $\uparrow$ \\
\midrule
Retrain LLM & 1 & 0.4202 & 0.4772 & 0.6740 \\
\midrule
GA & 2.38e-6 & 0.3955 & 0.4075 & 0.3944 \\
EGUP\_w/o(GA) & 1.39e-6 & 0.4086 & 0.4255 & 0.4030 \\
EGU(GA) & 1.12e-5 & 0.4083 & 0.4156 & 0.3989 \\
EGUP(GA) & 1.12e-5 & 0.4094 & 0.4261 & 0.4064 \\
\midrule
NPO & 6.73e-6 & 0.4051 & 0.4203 & 0.4105 \\
EGUP\_w/o(NPO) & 1.12e-5 & 0.3962 & 0.4323 & 0.4042 \\
EGU(NPO) & 6.73e-6 & 0.4039 & 0.4187 & 0.4035 \\
EGUP(NPO) & 4.02e-6 & 0.4027 & 0.4234 & 0.4070 \\
\midrule
GD & 1.39e-6 & 0.3480 & 0.4047 & 0.4160 \\
EGUP\_w/o(GD) & 6.73e-6 & 0.3487 & 0.4373 & 0.4288 \\
EGU(GD) & 2.38e-6 & 0.3919 & 0.4344 & 0.4534 \\
EGUP(GD) & 1.83e-5 & 0.3520 & 0.4327 & 0.4181 \\
\midrule
NPO+GD & 0.0878 & 0.3518 & 0.4158 & 0.3942 \\
EGUP\_w/o(NPO+GD) & 0.0061 & 0.3993 & 0.4323 & 0.4211 \\
EGU(NPO+GD) & 0.6284 & 0.3426 & 0.4105 & 0.4199 \\
EGUP(NPO+GD) & 0.7934 & 0.3474 & 0.4259 & 0.4147 \\
\bottomrule
\end{tabular}
}
\end{table}

\begin{table}[t]
\centering
\caption{PHI Results on TOFU-10\% benchmark.}
\label{tab:phi_tofu10}
\resizebox{0.7\linewidth}{!}{
\begin{tabular}{l|cc|cc}
\toprule
\multirow{2}{*}{\textbf{Method}} & \multicolumn{2}{c|}{\textbf{Forget Perf.}} & \multicolumn{2}{c}{\textbf{Retain Perf.}} \\
\cmidrule{2-5}
& F.Q. $\uparrow$ & R-L $\uparrow$ & M.U. $\uparrow$ & R-L $\uparrow$ \\
\midrule
Retrain LLM & 1 & 0.4251 & 0.4917 & 0.7706 \\
\midrule
GA & 8.02e-5 & 0.3557 & 0.3878 & 0.3716 \\
EGUP\_w/o(GA) & 3.77e-5 & 0.3685 & 0.4296 & 0.3805 \\
EGU(GA) & 1.16e-4 & 0.3373 & 0.4056 & 0.3440 \\
EGUP(GA) & 2.36e-4 & 0.3399 & 0.4106 & 0.3505 \\
\midrule
NPO & 1.22e-8 & 0.3196 & 0.4197 & 0.3423 \\
EGUP\_w/o(NPO) & 5.44e-8 & 0.3123 & 0.4214 & 0.3133 \\
EGU(NPO) & 3.77e-5 & 0.3581 & 0.4299 & 0.3834 \\
EGUP(NPO) & 3.77e-5 & 0.3718 & 0.4366 & 0.3892 \\
\midrule
GD & 1.40e-6 & 0.3738 & 0.4032 & 0.4033 \\
EGUP\_w/o(GD) & 7.69e-6 & 0.3431 & 0.4219 & 0.3876 \\
EGU(GD) & 2.17e-6 & 0.3582 & 0.4412 & 0.4053 \\
EGUP(GD) & 5.52e-5 & 0.3356 & 0.4363 & 0.3925 \\
\midrule
NPO+GD & 2.56e-5 & 0.4211 & 0.4200 & 0.4438 \\
EGUP\_w/o(NPO+GD) & 3.77e-5 & 0.4236 & 0.4265 & 0.4420 \\
EGU(NPO+GD) & 0.3958 & 0.3601 & 0.4209 & 0.4211 \\
EGUP(NPO+GD) & 0.3417 & 0.3509 & 0.4244 & 0.4089 \\
\bottomrule
\end{tabular}
}
\end{table}

\begin{table}[t]
\centering
\caption{LLAMA2 Results on TOFU-1\% benchmark.}
\label{tab:llama_tofu1}
\resizebox{0.7\linewidth}{!}{
\begin{tabular}{l|cc|cc}
\toprule
\textbf{Method} & \multicolumn{2}{c|}{\textbf{Forget Perf.}} & \multicolumn{2}{c}{\textbf{Retain Perf.}} \\
\cmidrule{2-5}
& F.Q. $\uparrow$ & R-L $\uparrow$ & M.U. $\uparrow$ & R-L $\uparrow$ \\
\midrule
Retrain LLM & 1 & 0.3934 & 0.6305 & 0.9956 \\
\midrule
GA & 0.4046 & 0.4183 & 0.5590 & 0.7553 \\
EGUP\_w/o(GA) & 0.4046 & 0.4183 & 0.5590 & 0.7553 \\
EGU(GA) & 0.9188 & 0.3719 & 0.5660 & 0.7762 \\
EGUP(GA) & 0.9188 & 0.3719 & 0.5660 & 0.7762 \\
\midrule
NPO & 0.7659 & 0.2905 & 0.5568 & 0.7187 \\
EGUP\_w/o(NPO) & 0.7659 & 0.2905 & 0.5568 & 0.7187 \\
EGU(NPO) & 0.9188 & 0.3084 & 0.5504 & 0.6873 \\
EGUP(NPO) & 0.9188 & 0.3682 & 0.5654 & 0.7733 \\
\midrule
GD & 0.2657 & 0.2298 & 0.5255 & 0.4506 \\
EGUP\_w/o(GD) & 0.2657 & 0.2298 & 0.5255 & 0.4506 \\
EGU(GD) & 0.7659 & 0.3391 & 0.5325 & 0.6814 \\
EGUP(GD) & 0.7659 & 0.3391 & 0.5325 & 0.6814 \\
\midrule
NPO+GD & 0.7659 & 0.3793 & 0.6243 & 0.7737 \\
EGUP\_w/o(NPO+GD) & 0.7659 & 0.2410 & 0.6270 & 0.7312 \\
EGU(NPO+GD) & 0.9188 & 0.2894 & 0.6294 & 0.7188 \\
EGUP(NPO+GD) & 0.9188 & 0.2417 & 0.6454 & 0.8059 \\
\bottomrule
\end{tabular}
}
\end{table}


\begin{table}[t]
\centering
\caption{LLAMA2 Results on TOFU-5\% benchmark.}
\label{tab:llama_tofu5}
\resizebox{0.7\linewidth}{!}{
\begin{tabular}{l|cc|cc}
\toprule
\textbf{Method} & \multicolumn{2}{c|}{\textbf{Forget Perf.}} & \multicolumn{2}{c}{\textbf{Retain Perf.}} \\
\cmidrule{2-5}
& F.Q. $\uparrow$ & R-L $\uparrow$ & M.U. $\uparrow$ & R-L $\uparrow$ \\
\midrule
Retrain LLM & 1 & 0.3935 & 0.6083 & 0.9955 \\
\midrule
GA & 1.12e-5 & 0.4418 & 0.4921 & 0.4726 \\
EGUP\_w/o(GA) & 1.12e-5 & 0.4429 & 0.4988 & 0.4824 \\
EGU(GA) & 2.96e-5 & 0.4832 & 0.5072 & 0.5126 \\
EGUP(GA) & 1.12e-5 & 0.4808 & 0.5106 & 0.5112 \\
\midrule
NPO & 2.61e-7 & 0.3885 & 0.5107 & 0.4625 \\
EGUP\_w/o(NPO) & 1.46e-7 & 0.5561 & 0.5346 & 0.6181 \\
EGU(NPO) & 1.12e-5 & 0.4501 & 0.5037 & 0.4827 \\
EGUP(NPO) & 1.12e-5 & 0.4504 & 0.5119 & 0.4912 \\
\midrule
GD & 1.87e-9 & 0.0355 & 0.6106 & 0.7600 \\
EGUP\_w/o(GD) & 2.61e-7 & 0.9068 & 0.6502 & 0.6470 \\
EGU(GD) & 2.38e-6 & 0.0428 & 0.6166 & 0.5918 \\
EGUP(GD) & 2.38e-6 & 0.0621 & 0.6213 & 0.5645 \\
\midrule
NPO+GD & 7.54e-5 & 0.4559 & 0.5905 & 0.5681 \\
EGUP\_w/o(NPO+GD) & 7.54e-5 & 0.4245 & 0.6024 & 0.7238 \\
EGU(NPO+GD) & 0.7126 & 0.1367 & 0.6060 & 0.7990 \\
EGUP(NPO+GD) & 0.9647 & 0.3022 & 0.6202 & 0.7926 \\
\bottomrule
\end{tabular}
}
\end{table}

\begin{table}[t]
\centering
\caption{LLAMA2 Results on TOFU-10\% benchmark.}
\label{tab:llama_10}
\resizebox{0.7\linewidth}{!}{
\begin{tabular}{l|cc|cc}
\toprule
\textbf{Method} & \multicolumn{2}{c|}{\textbf{Forget Perf.}} & \multicolumn{2}{c}{\textbf{Retain Perf.}} \\
\cmidrule{2-5}
 & F.Q. $\uparrow$ & R-L $\uparrow$ & M.U. $\uparrow$ & R-L $\uparrow$ \\
\midrule
Retrain LLM & 1 & 0.3969 & 0.6187 & 0.9950 \\
\midrule
GA & 2.89e-11 & 0.0577 & 0.0 & 0.0756 \\
EGUP\_w/o(GA) & 4.32e-9 & 0.0132 & 0.0 & 0.0241 \\
EGU(GA) & 5.73e-7 & 0.0018 & 0.0 & 0.0040 \\
EGUP(GA) & 1.16e-5 & 0.0018 & 0.0 & 0.0040 \\
\midrule
NPO & 2.55e-9 & 0.5009 & 0.4987 & 0.5082 \\
EGUP\_w/o(NPO) & 1.49e-9 & 0.5206 & 0.5219 & 0.5217 \\
EGU(NPO) & 1.74e-7 & 0.4987 & 0.5012 & 0.5102 \\
EGUP(NPO) & 2.17e-7 & 0.4987 & 0.5178 & 0.5176 \\
\midrule
GD & 0.0017 & 0.2481 & 0.5281 & 0.3440 \\
EGUP\_w/o(GD) & 0.0023 & 0.1761 & 0.5521 & 0.4227 \\
EGU(GD) & 0.0055 & 0.0643 & 0.6114 & 0.6359 \\
EGUP(GD) & 0.0043 & 0.0761 & 0.6189 & 0.6659 \\
\midrule
NPO+GD & 5.02e-10 & 0.4490 & 0.5608 & 0.4815 \\
EGUP\_w/o(NPO+GD) & 1.22e-8 & 0.4402 & 0.5791 & 0.4843 \\
EGU(NPO+GD) & 0.2926 & 0.2752 & 0.5715 & 0.4973 \\
EGUP(NPO+GD) & 0.3417 & 0.2735 & 0.6315 & 0.7266 \\
\bottomrule
\end{tabular}
}
\end{table}
\begin{table*}[htbp]
\centering
\caption{Performance of different EGUP methods under various $k$ values and forget ratios on PHI. MU: Model Utility. FQ: Forget Quality.}\label{tab:varyk_phi}
\begin{tabular}{llcccccc}
\toprule
\textbf{Forget Ratio} & \textbf{Method} 
& \multicolumn{2}{c}{$k=0.5$} 
& \multicolumn{2}{c}{$k=1$} 
& \multicolumn{2}{c}{$k=2$} \\
\cmidrule(r){3-4} \cmidrule(r){5-6} \cmidrule(r){7-8}
& & MU & FQ & MU & FQ & MU & FQ \\
\midrule

\multirow{4}{*}{1\%} 
 & EGUP (GA)       & 0.4143 & 0.9999 & 0.4225 & 0.9999 & 0.4010 & 0.9999 \\
 & EGUP (GD)       & 0.4481 & 0.9900 & 0.4502 & 0.9900 & 0.4412 & 0.9900 \\
 & EGUP (NPO)      & 0.4012 & 0.9900 & 0.4133 & 0.9900 & 0.4120 & 0.9900 \\
 & EGUP (NPO+GD)   & 0.4638 & 0.9188 & 0.4610 & 0.9900 & 0.4521 & 0.9900 \\
\midrule

\multirow{4}{*}{5\%} 
 & EGUP (GA)       & 0.4390 & 2.61e-7 & 0.4156 & 1.12e-5 & 0.4071 & 1.12e-5 \\
 & EGUP (GD)       & 0.4263 & 4.61e-7 & 0.4344 & 2.38e-6 & 0.4239 & 4.61e-7 \\
 & EGUP (NPO)      & 0.3949 & 2.96e-5 & 0.4187 & 6.73e-6 & 0.4107 & 4.02e-6 \\
 & EGUP (NPO+GD)   & 0.4099 & 0.6284 & 0.4105 & 0.6284 & 0.3989 & 0.4663 \\
\midrule

\multirow{4}{*}{10\%} 
 & EGUP (GA)       & 0.4350 & 3.33e-6 & 0.4056 & 1.16e-4 & 0.4100 & 5.07e-6 \\
 & EGUP (GD)       & 0.4412 & 2.17e-6 & 0.4248 & 2.17e-6 & 0.4126 & 2.56e-5 \\
 & EGUP (NPO)      & 0.4196 & 2.72e-5 & 0.4298 & 3.77e-5 & 0.4210 & 3.77e-5 \\
 & EGUP (NPO+GD)   & 0.4407 & 0.1212 & 0.4209 & 0.3958 & 0.4202 & 0.2926 \\
\bottomrule
\end{tabular}
\end{table*}

\begin{table*}[htbp]
\centering
\vspace{1em}
\caption{Performance of different EGUP methods under various $k$ values and forget ratios on LLAMA2. MU: Model Utility. FQ: Forget Quality.}\label{tab:varyk_llama}

\begin{tabular}{llcccccc}
\toprule
\textbf{Forget Ratio} & \textbf{Method} 
& \multicolumn{2}{c}{$k=0.5$} 
& \multicolumn{2}{c}{$k=1$} 
& \multicolumn{2}{c}{$k=2$} \\
\cmidrule(r){3-4} \cmidrule(r){5-6} \cmidrule(r){7-8}
& & MU & FQ & MU & FQ & MU & FQ \\
\midrule

\multirow{4}{*}{1\%} 
 & EGUP (GA)       & 0.5567 & 0.4046 & 0.5660 & 0.9188 & 0.5376 & 0.5786 \\
 & EGUP (GD)       & 0.5705 & 0.1650 & 0.5781 & 0.1650 & 0.5325 & 0.7659 \\
 & EGUP (NPO)      & 0.5618 & 0.4046 & 0.5504 & 0.9188 & 0.5665 & 0.2657 \\
 & EGUP (NPO+GD)   & 0.6223 & 0.9188 & 0.6294 & 0.9188 & 0.5942 & 0.7659 \\
\midrule

\multirow{4}{*}{5\%} 
 & EGUP (GA)       & 0.4921 & 2.96e-5 & 0.5185 & 1.39e-6 & 0.5072 & 2.96e-5 \\
 & EGUP (GD)       & 0.6408 & 3.60e-9 & 0.6092 & 0.0002 & 0.6166 & 2.38e-6 \\
 & EGUP (NPO)      & 0.4600 & 6.57e-8 & 0.5248 & 1.46e-7 & 0.5037 & 1.12e-5 \\
 & EGUP (NPO+GD)   & 0.6519 & 0.1421 & 0.6060 & 0.7126 & 0.5917 & 0.2705 \\
\midrule

\multirow{4}{*}{10\%} 
 & EGUP (GA)       & 0 & 8,78e-12 & 0 & 5.73e-7 & 0.2294 & 1.07e-13 \\
 & EGUP (GD)       & 0.6114 & 0.0055 & 0.6259 & 1.16e-4 & 0.6247 & 2.17e-4 \\
 & EGUP (NPO)      & 0.5829 & 7.38e-15 & 0.5012 & 1.74e-7 & 0.3728 & 1.49e-9 \\
 & EGUP (NPO+GD)   & 0.6065 & 3.77e-5 & 0.5384 & 0.1761 & 0.5715 & 0.2926 \\
\bottomrule
\end{tabular}
\end{table*}


\end{document}